\documentclass{article}

\PassOptionsToPackage{numbers, compress}{natbib}

\usepackage[preprint]{neurips_2026}

\usepackage[utf8]{inputenc} 
\usepackage[T1]{fontenc}    
\usepackage{hyperref}       
\hypersetup{hidelinks}
\usepackage{url}            
\usepackage{booktabs}       
\usepackage{amsfonts}       
\usepackage{nicefrac}       
\usepackage{microtype}      
\usepackage{xcolor}         

\usepackage{amsmath}
\usepackage{graphicx}
\usepackage{bm}

\newcommand{\vx}{\bm{x}}
\newcommand{\vp}{\bm{p}}

\newcommand{\vq}{\bm{q}}

\newcommand{\va}{\bm{a}}
\newcommand{\vb}{\bm{b}}
\newcommand{\vh}{\bm{h}}
\newcommand{\vy}{\bm{y}}
\newcommand{\vz}{\bm{z}}

\newcommand{\authornote}[1]{\textsuperscript{\rlap{#1}}}

\title{Predictive Statistics Shape Emergent World Representations of Grid Walkers}

\author{%
\begin{tabular}{ccc}
Sasha Brenner\authornote{1,2} & Thomas R. Knösche\authornote{1,*} & Nico Scherf\authornote{1,3,*} \\
\texttt{brenner@cbs.mpg.de} & \texttt{knoesche@cbs.mpg.de} & \texttt{nscherf@cbs.mpg.de}
\end{tabular} \\
\\
$^1$ Max Planck Institute for Human Cognitive and Brain Sciences, Leipzig, Germany \\
$^2$ Leipzig University, Germany \\
$^3$ ScaDS.AI, Dresden/Leipzig, Germany \\
$^*$ Equal contribution
}

\begin{document}

\maketitle

\begin{abstract}
Next-token predictors often appear to develop internal representations of the latent world and its rules. The probabilistic nature of these models suggests a deep connection between the structure of the world and the geometry of probability distributions. In order to understand this link more precisely, we use a minimal stochastic process as a controlled setting: constrained random walks on a two-dimensional lattice that must reach a fixed endpoint after a predetermined number of steps. Optimal prediction of this process solely depends on a sufficient vector determined by the walker's position relative to the target and the remaining time horizon; in other words, the probability distributions are parametrized by the world's grid geometry. We train decoder-only transformers and recurrent networks on prefixes sampled from the exact distribution of these walks and compare their hidden activations to sufficient statistics of prediction, by measuring alignment and linear readability across layers. We find that the transformer's computation factors into two stages: the first attention block extracts the sufficient statistic from the input, and later layers transform it into the next-step predictive geometry. Across constraint variants the post-attention representation is universal: a shared world-state of the lattice that can be read directly as a world model, traced to the predictive geometry of the data. Later layers then specialize it to each variant's next-step distribution. Recurrent networks reach the same Bayes-optimal loss but do not isolate this world-state as a separate stage, showing that the world-model geometry also depends on architecture. Although demonstrated in a toy system, the results suggest that the geometry of the predictive distribution is a useful lens on how neural networks internalize the structure of their data.
\end{abstract}

\section{Introduction}
Modern sequence models develop internal representations whose geometric organization often mirrors structure of the world they are trained on \cite{Li2022EmergentWorldRepresentationsOthello, Gurnee2023LanguageModelsRepresentSpaceTime, Ivanitskiy2023StructuredWorldRepresentationsMazes, Karvonen2024EmergentWorldModelsChess, shaiTransformersRepresentBelief2024,  baumgartnerCognitiveMaps2025, park2025iclr, karkada2026symmetrylanguagestatisticsshapes, wurgaft2026manifoldsteeringrevealsshared}.
These observations are broadly suggestive of an emergent ``world model'', but the connection between data and representation has mostly been drawn by direct comparison between world structure and representation structure.

A common thread runs through these observations: the geometry of internal representations tends to reflect the geometry of probability distributions defined over the data. Different lines of work have made this connection through different distributions. One line under \emph{computational mechanics} found evidence that the belief states of generalized hidden Markov models tend to be linearly recoverable from the residual stream of trained transformers and RNNs \cite{shaiTransformersRepresentBelief2024, piotrowskiConstrainedBeliefUpdates2025, riechersNeuralNetworksLeverage2025, shai2026transformerslearnfactoredrepresentations}. Zhang et al.~\cite{zhangWhatShouldEmbeddings2025} give a Bayesian account of the same pattern, arguing that autoregressive embeddings should encode sufficient statistics of the latent generating distribution needed for prediction. Karkada et al.~\cite{karkada2026symmetrylanguagestatisticsshapes} analytically connect symmetries of word co-occurrence statistics to the manifold geometry of learned embeddings; Wurgaft et al.~\cite{wurgaft2026manifoldsteeringrevealsshared} link the activation manifold of a language model to the manifold structure of its next-token probability distribution.

We study the relationship between probability distributions and neural representations in a simple, flexible setting: constrained random walkers on a 2D square lattice. By varying the constraints within the same fixed grid world, we define a tractable family of prediction problems where the task is cleanly separated from the world. In all cases the spatiotemporal variables $(t, x, y)$ are a sufficient statistic for prediction of the full future---knowledge of the world state is sufficient for all variants. However, world-state knowledge is not \emph{necessary} for a next-step prediction task, which is the relevant goal for autoregressive sequence models. Indeed, their next-step log-probabilities are linearly obtainable from an associated constraint-dependent effective field $\vq_t(x, y) \in \mathbb{R}^3$, which is strictly less informative than the world state. Thus, these systems offer a controlled setting where we can ask not only whether neural networks trained on sequences of walker-steps learn to represent the underlying world, but also the precise way in which this world representation relates to the full stochastic process in question.

We train decoder-only transformers on sequences of these walkers' steps and show that a universal embedding consistently emerges across constraint variants by a simple mechanistic implementation. 
More generally, we ask whether the sufficient statistic for prediction is linearly related to the networks' residual stream---both as a quantity decodable from activations and as one that linearly spans them---, and if so, which parametrization of this statistic is represented. The answer depends on the layer: immediately after attention, the residual stream cleanly exposes a coordinate system of the world---sufficient for the full future---, while near the output, the representation is instead closer to the effective field $\vq$. The post-attention geometry is strikingly similar across walker variants, with cross-variant lCKA reaching $0.95$--$0.97$; the MLP then breaks this universality by specializing to each variant's predictive distribution---cross-variant similarity there drops sharply. Finally, we train recurrent neural networks (RNNs) on the same tasks and show that they solve them equally well and must track the same world-state, yet do not expose it as the separate, explicit coordinate stage the transformer builds; this geometric separation is thus shaped by architecture rather than dictated by prediction. Although our setting is deliberately simple, the analysis suggests a general mechanism by which world representations might sometimes emerge from prediction: layers may distill relevant statistics from the input---including world states---and later layers transform them into the task-specific predictive distribution.

The remainder of this paper is organized as follows. Section~\ref{sec:related_work} reviews prior work on emergent world models and on the geometry of predictive representations. Section~\ref{sec:methods} introduces the constrained grid-walker process and the analytical quantities used for training and evaluation. 
Section~\ref{sec:results} reports our representational findings. 
Section~\ref{sec:limitations} discusses limitations, and Section~\ref{sec:conclusions} concludes.

\section{Related work}
\label{sec:related_work}
\subsection{Emergent world models}
Several studies have shown that next-token predictors develop internal representations that function as world models. Early evidence in a controlled setting came from a GPT model trained on Othello move sequences, which developed an emergent internal representation of the board state that could be used to control model outputs via interventions \cite{Li2022EmergentWorldRepresentationsOthello, nandaEmergentLinearRepresentations2023}. Similar findings were subsequently reported for chess-playing language models \cite{Karvonen2024EmergentWorldModelsChess}. In large-scale LLMs, representations have been shown to encode linear structure for space and time \cite{Gurnee2023LanguageModelsRepresentSpaceTime}. Complementary evidence from spatial and graph-structured prediction tasks, studied under both standard training and in-context learning, shows that transformers develop structured internal representations of topology, paths, and space \cite{Ivanitskiy2023StructuredWorldRepresentationsMazes,baumgartnerCognitiveMaps2025,park2025iclr}. Across these settings, however, the complexity of the tasks makes it difficult to say precisely which statistical structure of the prediction problem gives rise to the learned geometry.

\subsection{The geometry of predictive representations}
\label{sec:simplex_related_work}
A growing body of work argues that the geometry of internal representations in trained sequence models is shaped by the geometry of probability distributions defined over the data. The relevant distribution, however, differs across lines of work. Computational mechanics studies the predictive distribution $\Pr(\text{future}\mid\text{past})$ and its sufficient statistic: Shai et al.~\cite{shaiTransformersRepresentBelief2024} showed that small transformers trained on edge-emitting hidden Markov models develop linear mappings to the belief-state simplex; Piotrowski et al.~\cite{piotrowskiConstrainedBeliefUpdates2025} subsequently proposed and tested a mechanistic implementation of belief updating in self-attention; and Riechers et al.~\cite{riechersNeuralNetworksLeverage2025} extended the analysis to RNNs and a broader class of generative models, all in stationary and ergodic settings. Independently, Zhang et al.~\cite{zhangWhatShouldEmbeddings2025} argue from a Bayesian perspective that autoregressive embeddings should encode sufficient statistics for prediction. A separate line studies other distributions over the training data: Karkada et al.~\cite{karkada2026symmetrylanguagestatisticsshapes} show that symmetries of word co-occurrence statistics analytically govern the manifold geometry of trained embeddings, and Wurgaft et al.~\cite{wurgaft2026manifoldsteeringrevealsshared} connect activation geometry to the geometry of model output behavior in larger language and multimodal models. Across these threads, the recurring observation is the same: internal representation geometry reflects the geometry of a probability distribution over the data.

Our setting brings both perspectives together. The constrained walker has a simple, perturbable world---a 2D lattice with goals and obstacles that can be modified independently---so its world structure is naturally studied in the spirit of \cite{Gurnee2023LanguageModelsRepresentSpaceTime, karkada2026symmetrylanguagestatisticsshapes, wurgaft2026manifoldsteeringrevealsshared}. At the same time, the process admits a fully analytic probabilistic description: the predictive distribution and its sufficient statistic are explicit objects, and the perspective of \cite{shaiTransformersRepresentBelief2024, riechersNeuralNetworksLeverage2025, zhangWhatShouldEmbeddings2025} applies without approximation. The shared world across walker variants further lets us examine how a representation is reshaped layer by layer to apply variant-specific constraints. Moreover, since the grid walker is a non-stationary, non-ergodic process, our results extend the findings of \cite{shaiTransformersRepresentBelief2024, piotrowskiConstrainedBeliefUpdates2025, riechersNeuralNetworksLeverage2025} to a qualitatively different class of processes.

\section{Methods}
\label{sec:methods}
We work in a setting where the ground truth is available exactly: constrained random walkers on a 2-dimensional square lattice; \emph{walkers} for short. For each walker configuration, we compute the full joint probability distribution of the process, train decoder-only transformers and recurrent networks from scratch on prefix sequence samples drawn from these distributions with a next-token cross-entropy loss, and evaluate learned hidden states by fitting linear probes to analytically derived predictive vectors across layers. 
The reproduction code for data generation, model training, representational analyses, and figure generation is available at \url{https://github.com/allshaks/grid-walkers/tree/master}.

\begin{figure}[t]
  \centering
  \includegraphics[width=0.7\linewidth]{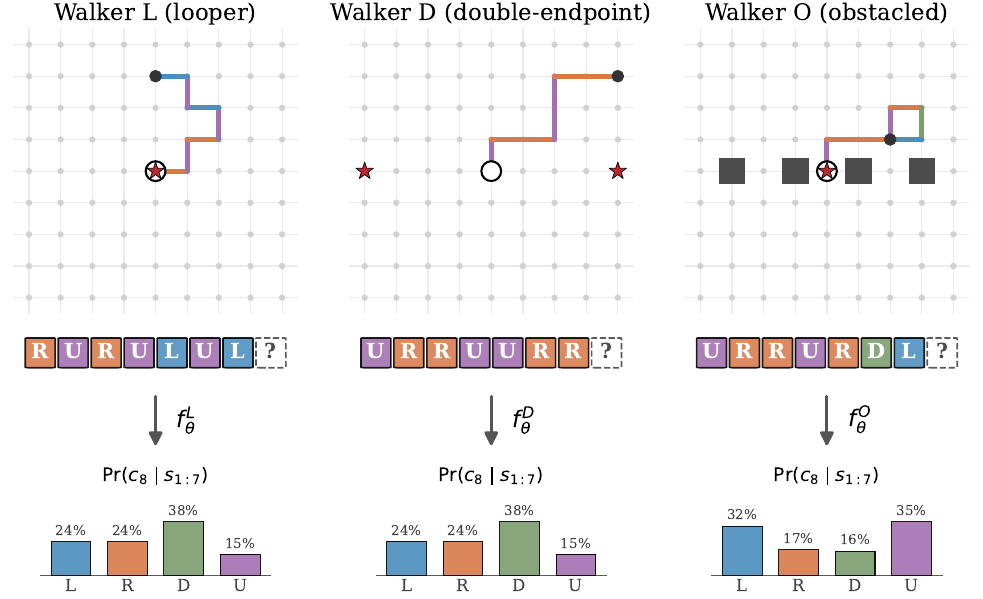}
  \caption{The three walker variants and the prediction task. Each column shows one walker (left to right: \textbf{walker L} (looper), \textbf{walker D} (double-endpoint), \textbf{walker O} (obstacled)) and reads top to bottom: a length-7 prefix on the lattice, one color per move type, final position as a black dot; the emitted token sequence, with a ``?'' for the unknown next move; and the next-step distribution $\Pr(c_8 \mid s_{1:7})$ produced by the per-walker model $f^{L}_\theta, f^{D}_\theta, f^{O}_\theta$. Endpoints are red stars, obstacles dark squares, the start a white circle.}
  \label{fig:walker_schematics}
\end{figure}

\begin{figure}[t]
  \centering
  \includegraphics[width=0.85\linewidth]{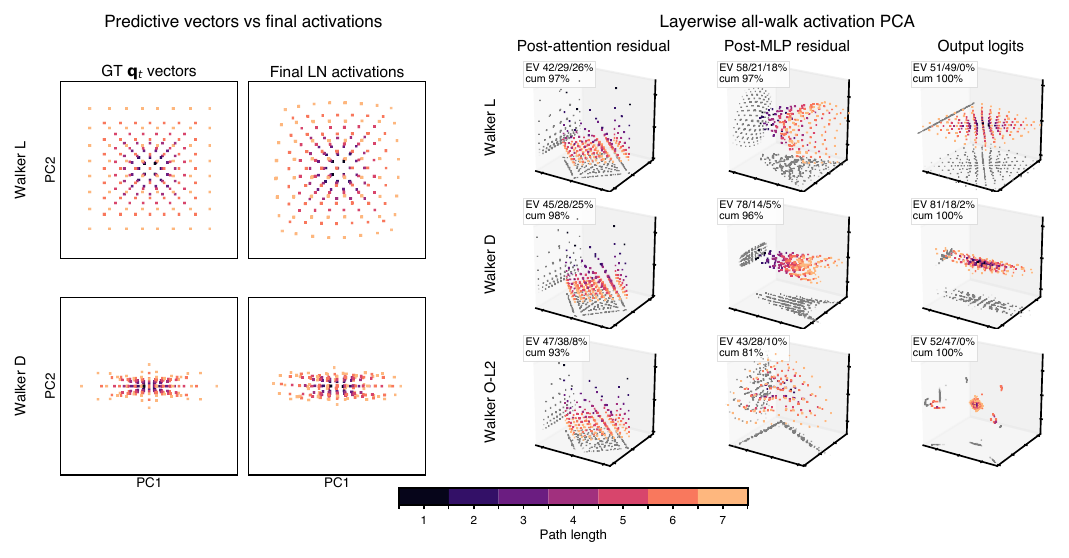}
  \caption{Activation geometry across targets and layers. Left: Last-LayerNorm activations (right column of each pair) closely resemble the ground-truth next-step targets (left column) for walker~L (top) and walker~D (bottom). For both walkers the target is the effective field $\vq_t$ of Equation~\ref{eq:eff_field}; for walker~L $\vq_t$ is effectively 2D since $q_{uv}=0$, whereas for walker~D it has all three components nonzero. Activations are Procrustes-aligned to the ground-truth target coordinates. Right: 3D PCA views of all valid walk-prefix activations across post-attention residual, post-MLP residual, and output logits for walkers L, D, and O-L2. Each point is a distinct path on the lattice, colored by path length with a shared colorbar.}
  \label{fig:gt_acts_comparison}
\end{figure}

\subsection{The grid walker}
\label{sec:grid_theory}

A grid walker takes discrete steps on $\mathbb{Z}^2$ from the origin with a fixed time horizon $T$, choosing at each timestep $t$ one of four moves $c_t \in \{L, R, D, U\}$ with displacements $\Delta_L=(-1,0)$, $\Delta_R=(1,0)$, $\Delta_D=(0,-1)$, $\Delta_U=(0,1)$. We write $s_{a:b}:=(c_a,\ldots,c_b)$ for a sequence of moves and abbreviate the length-$t$ prefix as $s_t:=s_{1:t}$; its corresponding position is $\vx_t = \sum_{t'=1}^t \Delta_{c_{t'}}$.

\paragraph{General setting.}
The walker is parameterized by an endpoint set $\mathcal{P}\subset\mathbb{Z}^2$ and a (possibly empty) obstacle set $\mathcal{O}\subset\mathbb{Z}^2$. Samples are drawn by first picking $\vp\in\mathcal{P}$ uniformly and then sampling uniformly among length-$T$ paths from the origin to $\vp$ that avoid $\mathcal{O}$. The pair $(t,\vx_t)$ is a sufficient statistic for prediction: conditioned on it, the distribution over future continuations no longer depends on the detailed prefix. This is the predictive-state sense in which state captures the information from the past relevant to predicting the future \cite{shaliziComputationalMechanicsPatternPrediction2002,littmanPredictiveRepresentationsState2001}. Note that partial walks of length $t<T$ are not uniformly distributed, 
with probability proportional to the number of valid completions; equivalently, each endpoint-conditioned process is a finite-horizon Doob $h$-transform of the simple random walk \cite{doobConditionalBrownianMotion1957,levinMarkovChainsMixing2017} (Appendix~\ref{app:distrib_comp}).

It is convenient to work in \emph{diagonal coordinates} $u_t := x_t + y_t,\ v_t := x_t - y_t$, in which the four lattice moves correspond to the four sign choices for the diagonal components $(\Delta u, \Delta v)$ of their displacement $\Delta$: $\Delta_R\mapsto(+1,+1)$, $\Delta_L\mapsto(-1,-1)$, $\Delta_U\mapsto(+1,-1)$, $\Delta_D\mapsto(-1,+1)$. Each step shifts $u_t$ and $v_t$ independently by $\pm 1$, so the unconstrained walker is equivalent to two identically independent 1D walks in $(u, v)$.

Since the four-element step basis $\{1, \Delta u, \Delta v, \Delta u\,\Delta v\}$ is complete on $\{\pm 1\}^2$, every next-step distribution admits the exact decomposition
\begin{equation}
  \Pr(\Delta \mid s_t;\mathcal{P}, \mathcal{O}) \;=\; \tfrac{1}{4}\bigl(1 + \langle\Delta u\rangle\,\Delta u + \langle\Delta v\rangle\,\Delta v + \langle\Delta u\,\Delta v\rangle\,\Delta u\,\Delta v\bigr)
  \label{eq:central_identity}
\end{equation}

where $\langle\cdot\rangle$ denotes the expectation under the next-step distribution and $\Delta \in \{\Delta_L, \Delta_R, \Delta_U, \Delta_D\}$ is a generic next-step displacement. In logit space, the mean-centered log-probabilities also admit the decomposition
\begin{equation}
  \ell_\Delta - \bar\ell = -q_u\,\Delta u - q_v\,\Delta v - q_{uv}\,\Delta u\,\Delta v,
  \label{eq:eff_field}
\end{equation}

where $\bar\ell = \tfrac{1}{4}\sum_\Delta \ell_\Delta$ is the mean logit and the \emph{effective field} $\vq := (q_u, q_v, q_{uv})$ collects the three coefficients. Exponentiating Equation~\ref{eq:eff_field} and normalizing recovers the Softmax form
\begin{equation}
    \Pr(\Delta\mid s_t;\mathcal{P}, \mathcal{O}) = \frac{e^{-\vq_t\cdot\boldsymbol{\Delta}_\star}}{\sum_{\Delta'} e^{-\vq_t\cdot\boldsymbol{\Delta}'_\star}},\qquad
    \boldsymbol{\Delta}_\star := (\Delta u,\, \Delta v,\, \Delta u\,\Delta v),
    \label{eq:next_step_probs}
\end{equation}
where $\boldsymbol{\Delta}_\star$ is the three-coordinate representation of step $\Delta$ in the natural logit basis.

When there are no obstacles ($\mathcal{O} = \emptyset$), the next-step probability given an endpoint $\vp$ factorizes as
\begin{equation}
  \Pr(\Delta \mid s_t;\vp) = \tfrac{1}{4}(1 + \mu_p\,\Delta u)(1 + \nu_p\,\Delta v),\qquad
  \mu_p := \frac{p_u - u_t}{\tau},\qquad
  \nu_p := \frac{p_v - v_t}{\tau},
  \label{eq:next_step_probs_product}
\end{equation}
where $p_u = p_x + p_y,\ p_v = p_x - p_y$ and $\tau = T - t$ is the remaining number of steps. The general multi-endpoint distribution is the posterior mixture
\begin{equation}
    \Pr(\Delta \mid s_t;\mathcal{P}) = \sum_{\vp \in \mathcal{P}} w_t(\vp\mid\vx_t)\, \Pr(\Delta\mid s_t;\vp),
    \label{eq:multi_endpoint_mix}
\end{equation}
where $w_t(\vp\mid\vx_t)$ is the Bayesian posterior over endpoints (Appendix~\ref{app:distrib_comp}).

\paragraph{Single-endpoint variant (Walker L).}
For $\mathcal{O}=\emptyset$ and $|\mathcal{P}|=1$ Equation \ref{eq:next_step_probs_product} holds, so the next-step moments coincide with the drifts: $\langle\Delta u\rangle = \mu_p$, $\langle\Delta v\rangle = \nu_p$, $\langle\Delta u\,\Delta v\rangle = \mu_p\nu_p$. In particular $\langle\Delta u\,\Delta v\rangle = \langle\Delta u\rangle\langle\Delta v\rangle$, so the third logit coefficient in Equation \ref{eq:eff_field} vanishes: $q_{uv} = 0$. The effective field then collapses to 2D, $\vq_t = (q_u, q_v, 0)$, with closed-form artanh expressions for $q_u, q_v$ in $(\mu_p, \nu_p)$ (Appendix~\ref{app:gaussian_counts}).

\paragraph{Double-endpoint variant (Walker D).}
For $\mathcal{O}=\emptyset$ and $|\mathcal{P}|=2$ the endpoint posterior is non-trivial, and the next-step moments admit an explicit bridge-field interpretation: $\langle\Delta u\rangle = \langle\mu_t\rangle_\mathcal{P}$, $\langle\Delta v\rangle = \langle\nu_t\rangle_\mathcal{P}$, $\langle\Delta u\,\Delta v\rangle = \langle\mu_t\nu_t\rangle_{\mathcal{P}}$, where $\langle\cdot\rangle_\mathcal{P}$ denotes the posterior average of the bridge fields over $w_t(\vp\mid\vx_t)$. Let $w_\pm$ denote the posterior weights, $w_+ + w_- = 1$, and $s := w_+ - w_- \in [-1, 1]$ the belief asymmetry. The posterior covariance then satisfies
\begin{equation*}
  \langle\mu\nu\rangle - \langle\mu\rangle\langle\nu\rangle \;=\; \frac{16(1-s^2)}{\tau^2},
\end{equation*}
nonzero whenever $|s| < 1$. The factorization of Equation~\ref{eq:next_step_probs_product} is then broken: $q_{uv}\neq 0$, no 2D reduction exists, and all three components of $\vq_t$ carry independent information about position and belief. 

\paragraph{Obstacled variant (Walker O).}
For nonempty $\mathcal{O}$, translation invariance is broken and Equation~\ref{eq:next_step_probs_product} no longer applies: the bridge to $\vp$ avoiding $\mathcal{O}$ has no closed-form factorization in $(u,v)$ (Appendix~\ref{app:distrib_comp}). Nevertheless, the central identity of Equation~\ref{eq:central_identity} and the effective-field expression of Equation~\ref{eq:eff_field} continue to apply. We write the effective field as a correction to the free field, $\vq^O_t = \vq^L_t + \Delta\vq^O_t$, where $\vq^L_t$ is the obstacle-free reference field to the same endpoint and $\Delta\vq^O_t$ is the obstacle-induced perturbation (Appendix~\ref{app:eff_field}). We introduce this variant for two reasons: to explicitly break some symmetry in order to challenge the networks with a slightly harder learning task, but also to construct a controlled test of representational fidelity: by placing obstacles such that several distinct positions share an identical next-step distribution while differing in their longer-horizon futures, we can ask whether the network maintains the position distinction internally even though the immediate next-step distribution does not distinguish these states. This is the spatial analog of the RRXOR-style degeneracy test introduced for hidden Markov models in \cite{shaiTransformersRepresentBelief2024}.

\subsection{Data generation}
\label{sec:data_gen}
We train models on length-$K$ contexts drawn from the marginal distribution over prefixes induced by the full conditioned process (Equation~\ref{eq:seq_marginal} in Appendix~\ref{app:distrib_comp}), computed via dynamic programming. We keep a fixed context length $K=8$\footnote{Note that shorter prefixes with $t<K$ still contribute to the next-token loss.}, small enough to preserve tractability.

For our main experiments we focus on three concrete walker configurations that instantiate the three settings introduced above (Figure~\ref{fig:walker_schematics}):
\begin{itemize}
    \item \textbf{Walker L} (looper) — endpoint $\vp = (0,0)$, horizon $T = 20$.
    \item \textbf{Walker D} (double-endpoint) — endpoint set $\mathcal{P} = \{(-4,0), (4,0)\}$, horizon $T = 20$.
    \item \textbf{Walker O} (obstacled) — endpoint $\vp = (0,0)$, horizon $T = 20$, with obstacles at $\{(\pm 1, 0), (\pm 3, 0)\}$. The obstacles are placed so that the next-step distribution at $(0,0)$, $(2,0)$, and $(-2,0)$ is identical (only the up/down displacements are available), even though the longer-horizon futures of these positions differ, the former being the actual endpoint.
\end{itemize}
We use \textbf{walker L}, \textbf{walker D}, and \textbf{walker O} as shorthand for these variants throughout the paper.

\subsection{Neural Models}

\paragraph{Architectures.}
We train decoder-only transformers \cite{vaswaniAttentionAllYou2017} from scratch on each walker, sweeping over architectures with $L\in\{1,2\}$ layers, $H\in\{1,2\}$ attention heads, and head dimension $d_{\rm head}\in\{32,64,128\}$. Unless otherwise noted, main-text figures show Walker~L and Walker~D at the single-layer, single-head configuration ($L=1, H=1$), and Walker~O at $L=1$ and $L=2$ (both with $H=2$), which we abbreviate Walker~O-1L and Walker~O-2L; the remaining architectures in the sweep are reported in the appendix. For the architecture comparison, we additionally train recurrent baselines (LSTM and GRU cells, $L\in\{1,2\}$ layers, hidden sizes $H\in\{16,\ldots,256\}$) on the same data; see Appendix~\ref{app:rnn_comparison}.

\paragraph{Sampling and training.}
Training samples are drawn in minibatches with replacement from the marginal distribution. The objective is next-token cross-entropy loss over all context positions, averaged over the minibatch: the term at position $t$ predicts $c_{t+1}$ given the prefix $s_t$. Because there is no beginning-of-sequence token, the first term predicts $c_2$ from $c_1$. All models are optimized with AdamW; full hyperparameters are reported in Appendix~\ref{app:hyperparams}.

\paragraph{Validation.}
Validation uses exact expected cross-entropy loss under the ground-truth distribution. The per-token loss is calculated as in training, but instead of the sample mean of a minibatch, we compute a weighted average over the entire set of possible prefixes at every validation round, where each partial sequence's per-token loss is weighted by its exact ground-truth probability. Importantly, this validation paradigm does not measure generalization, but controls for the global optimization problem. Final validation losses of the trained models are shown in Appendix~\ref{app:training_curves}, Table~\ref{tab:walkers}, with respect to a baseline theoretical loss given by the ground-truth conditional entropy of an observation given a history.

\section{Results and discussion}
\label{sec:results}

Across all three walker variants, the trained transformers reach the Bayes-optimal validation loss. We find that their internal representations closely track sufficient statistics for prediction at two distinct stages of the residual stream: one for the full future at post-attention, and another for the next-step distribution at LN-final, as we will show in the rest of this section. In Figure~\ref{fig:gt_acts_comparison}, the top two principal components of the final-LayerNorm activations resemble the corresponding ground-truth next-step target $\vq_t$: for Walker~L this target is effectively 2D, while for Walker~D the plot shows the top two PCs of the full 3D $\vq_t$. The rightmost panels of the figure, on the other hand, hint that the geometry at the post-attention residual is largely shared across variants, while the geometry at the post-MLP residual is more variant-specific. The next subsections develop this observation.

\subsection{Transformers linearly represent sufficient statistics}
\label{sec:what_is_encoded}
\label{sec:alignment_results}

\begin{figure}[t]
  \centering
  \includegraphics[width=\linewidth]{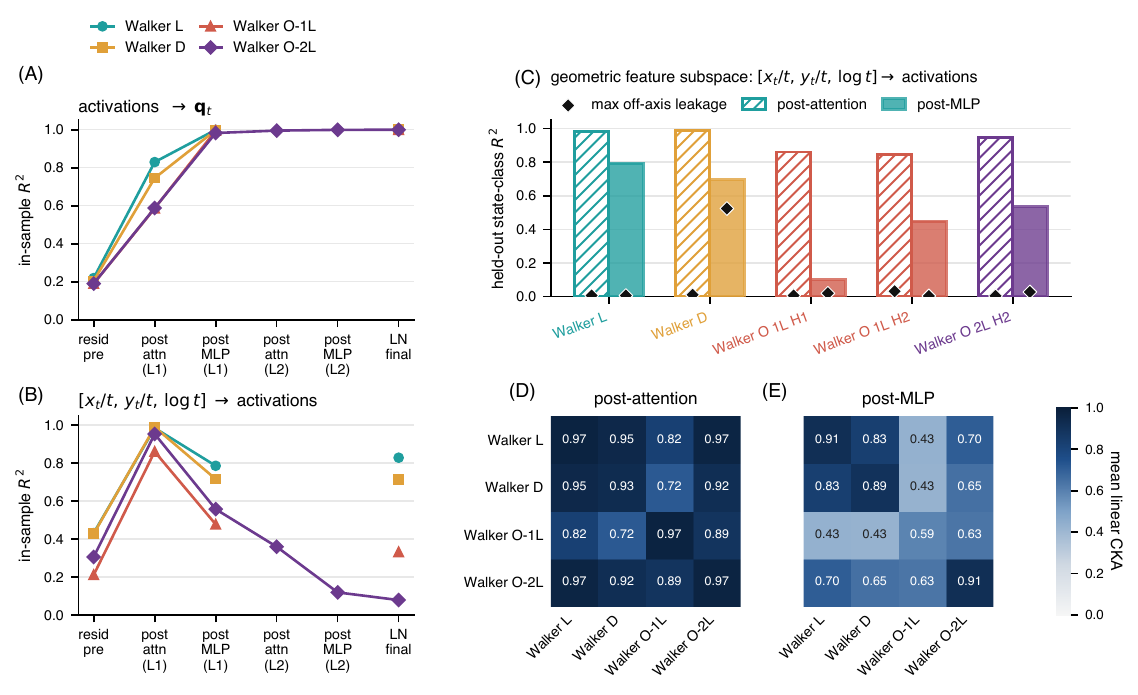}
  \caption{Linear probe alignment and cross-variant similarity ($d_{\rm head}=64$). \textbf{(A,B)} All-walk in-sample affine probes on the residual stream, fit in \emph{opposite} directions at each hook; gaps mark second-layer hooks, absent in one-layer models. \textbf{(A) Decoding direction} ($\text{activations}\to\vq$): joint $R^2$ of an affine probe predicting the 3D effective field from the activation. \textbf{(B) Reverse direction} ($[x_t/t, y_t/t, \log t]\to\text{activations}$): joint $R^2$ of an affine probe predicting the activation from the 3D geometric features. High $R^2$ here means the activation lies in the affine span of the features. \textbf{(C)} Held-out state-class $R^2$ of a linear probe onto the 3D geometric features, comparing post-attention (hatched) and post-MLP (solid) bars across walker variants and architectures. Diamonds show the maximum off-diagonal feature-axis leakage. \textbf{(D,E)} Pairwise linear CKA between state-mean activations at the post-attention (D) and post-MLP (E) hooks, computed on the common support intersection of states valid across the selected runs. Each entry averages the pairwise CKA over all distinct run pairs across the two groups; diagonal entries average over distinct same-group run pairs and thus measure within-variant reproducibility. Before the MLP, representations are nearly identical across variants (CKA $\geq 0.72$); after the MLP, cross-variant similarity drops sharply, with the structurally most distinct Walker~O-1L diverging the most.}
  \label{fig:r2_combined}
\end{figure}

We probe the residual stream at two stages and find that each linearly encodes a sufficient statistic for prediction, in line with the linear representation hypothesis \cite{park2023linear}. At LN-final this is almost by construction: the next-step logits are a linear readout of the activations and $\vq_t$ is a fixed linear transform of the logits, so an affine probe onto $\vq_t$ (Figure~\ref{fig:r2_combined}A) saturates across walkers and head dimensionality (Appendix~\ref{app:additional_diagnostics}, Figure~\ref{fig:activation_to_feature_r2_all_dheads}). At the post-attention residual, $\vq_t$ is not yet cleanly recoverable—so what is encoded there?

In fact the post-attention residual encodes a strictly richer statistic: the geometric features $[x_t/t,\, y_t/t,\, \log t]$, sufficient for the full future rather than only the next step. An affine probe recovers them cleanly across all walker variants (Figure~\ref{fig:activation_to_feature_r2_all_dheads}). But decodability only shows the features are \emph{present} in the activation, not that they are what it is built from. To test the stronger claim we fit the opposite mapping, from the three features into the high-dimensional activation (Figure~\ref{fig:r2_combined}B), reaching $R^2 = 0.86$--$0.99$. Since $d_{\rm features}=3 \ll d_{\rm head}\in\{32,64,128\}$, such a high $R^2$ means that the geometric statistic does not merely sit in the post-attention residual, it essentially spans it. An analogous reverse fit appears in \cite{zhangWhatShouldEmbeddings2025}, who nonlinearly reconstruct embeddings from a model's sufficient statistics; our affine probe sharpens this to a statement about the activations' linear span.

Strikingly, the features persist even where the next step cannot distinguish them: for Walker~O, positions $(-2,0)$, $(0,0)$, and $(2,0)$ share an identical next-step distribution, yet the geometric features—which separate these positions—still explain the post-attention activations well, echoing \cite{shaiTransformersRepresentBelief2024}, where belief states were linearly readable despite similarly degenerate next-step states. One might read this retained positional information as incomplete compression—but that sees only half the picture. In a single, cheap step the first attention module has already discarded the exact ordering of moves, a large and ubiquitous source of variation in the data that is irrelevant to prediction. Against that, keeping some strictly unnecessary position is a minor residue: compression is not finished, but most of it is already done, almost for free. We turn next to the mechanism that does it.

\subsection{Attention constructs an architecture-specific coordinate system}
\label{sec:first_attention_mechanism}

The mechanism that encodes the sufficient statistic $[x_t/t, y_t/t, \log t]$ at the first self-attention module, common to all walker variants, combines three properties of the trained attention head (full derivation in Appendix~\ref{app:attention_mechanism}). Figure~\ref{fig:first_layer_encoding} shows a representative Walker~L run; the same uniform-attention, additive value-write pattern was verified across the trained walker variants and architecture sweep:

\begin{enumerate}
  \item \textbf{Uniform averaging.} The attention head averages value-write contributions approximately uniformly over prefix positions $1,\ldots,t$ (Figure \ref{fig:first_layer_encoding}A).
  \item \textbf{Additive value writes.} Each value-write vector decomposes into a token-specific direction code $\va_{c_i}$ and a position-specific time code $\vp_i$, with negligible cross-coupling between the two.
  \item \textbf{Token and positional codes.} The direction codes arrange the four lattice moves at opposite poles of a 2D subspace (Figure~\ref{fig:first_layer_encoding}B), so their average is a linear function of the mean displacement:
  \begin{equation}
  \frac{1}{t}\sum_{i=1}^t \va_{c_i}
  \;\simeq\;
  A \frac{1}{t}\sum_{i=1}^t \Delta_{c_i}
  \;=\;
  A\,\frac{\vx_t}{t},
  \label{eq:compass_avg}
  \end{equation}
  for $A = \left[ \begin{matrix} \va_R^\top \\ \va_U^\top \end{matrix}\right]$. Averaging the position codes gives the time code $\vh_t := \frac{1}{t}\sum_{i=1}^t \vp_i$, which depends only on $t$.
\end{enumerate}

The attention output therefore decomposes into a 2D spatial coordinate $\vx_t/t$ and a deterministic time code $\vh_t$—the right panel of Figure~\ref{fig:gt_acts_comparison} shows how its activations reflect this geometry. As a compact scalar proxy, $\log t$ fits $\vh_t$ well empirically, though $\vh_t$ is not strictly one-dimensional (Figure~\ref{fig:first_layer_encoding}C): a second PCA component of the empirical time centroids captures approximately $6\%$ of their variance, so $\log t$ approximates only the dominant direction. Interestingly, a very similar mechanism to the one presented here has been described in \cite{bhattamishraAbilityLimitations2020} during a constructive analysis of transformers on counter languages.

\begin{figure}[t]
  \centering
  \includegraphics[width=\linewidth]{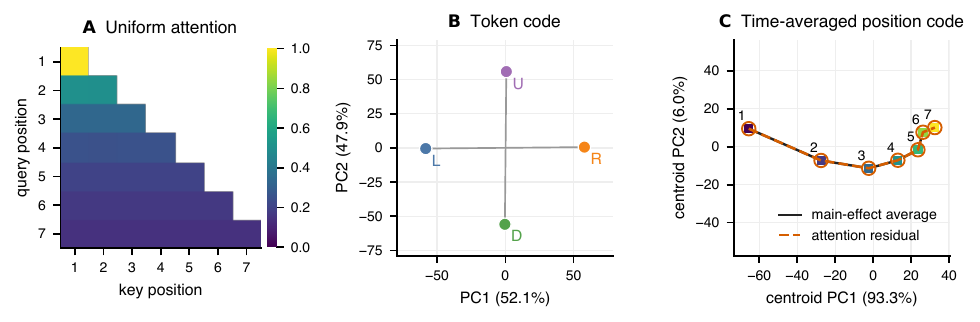}
  \caption{First-layer mechanism in a representative trained Walker~L transformer ($L=1,H=1,d_{\rm head}=128$). \textbf{(A)} The single attention head is nearly uniform over the causal prefix. \textbf{(B)} The token component of the value-write vectors forms a two-dimensional compass code for the four lattice moves. \textbf{(C)} The residual after removing the best affine $x_t/t,y_t/t$ component from the empirical attention output (orange dashed) is almost exactly the time average of the position component of the value-write vectors (black). Panel~C is plotted in the PCA basis of the seven empirical residual centroids.}
  \label{fig:first_layer_encoding}
\end{figure}

\subsection{All walker variants share common post-attention geometry}
\label{sec:universality}

Two analyses confirm that the post-attention representation is shared across walker variants. First, the three geometric axes generalize beyond fitted states: a probe evaluated on held-out state-mean activations keeps post-attention $R^2$ in $0.85$–$0.99$ across all variants and architectures, while post-MLP $R^2$ drops sharply, most for Walker~O (Figure~\ref{fig:r2_combined}C; Table~\ref{tab:feature_r2_arch} and Figure~\ref{fig:features_grid}, appendix). The small post-attention leakage in Figure~\ref{fig:r2_combined}C further shows the three axes are nearly orthogonal—each capturing its own coordinate with little cross-contamination—confirming the additive value-writes of Section~\ref{sec:first_attention_mechanism}.

Second, these representations are aligned across networks trained on different variants. Pairwise linear CKA \cite{kornblithSimilarityNeuralNetwork2019} between state-mean activations (Figure~\ref{fig:r2_combined}D,E) reaches $0.95$–$0.97$ between the well-trained variants before the MLP—indistinguishable from the within-variant diagonal—and stays above $0.72$ even for Walker~O-1L, despite the three processes having different endpoints and next-step distributions. After the MLP, cross-variant similarity drops sharply: Walker~L vs Walker~O-1L falls from $0.82$ to $0.43$ (full matrix in Figure~\ref{fig:cka_full}, appendix).

\subsection{Post-attention representations specialize downstream}

\begin{figure}[t]
  \centering
  \includegraphics[width=\linewidth]{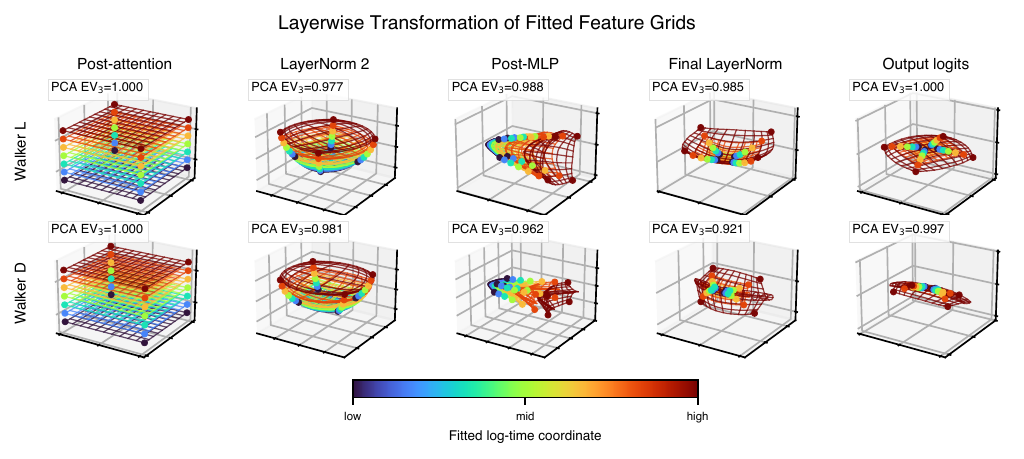}
  \caption{Layerwise transformation of synthetic feature grids. We construct grids spanning the actual datapoints at the post-attention residual stream of Walker~L (top row) and Walker~D (bottom row), with points colored by the fitted $\log t$ coordinate (low to high, see colorbar). Each subsequent column shows the image of these grids at LayerNorm 2, post-MLP, the final LayerNorm, and the output logits. The annotated PCA EV$_3$ is the fraction of variance captured by the top three principal components of the activations at each stage.}
  \label{fig:predictive_transform}
\end{figure}

To trace where the shared geometry described in the previous sections goes downstream, Figure~\ref{fig:predictive_transform} shows the layerwise fate of synthetic grids drawn in the fitted post-attention subspace $[x_t/t,\, y_t/t,\, \log t]$ (justified by the high $R^2$ there). The LayerNorms and MLP progressively bend these grids into walker-specific output-logit clouds that stay low-dimensional throughout: at least $91\%$ of output-logit variance lies in a 2D subspace across the full sweep (Figure~\ref{fig:output_logit_pca_dim}, appendix). Notably, the loss leaves a free constant shift of the logits unconstrained, yet the network does not exploit this freedom to inflate the output geometry.

\subsection{RNNs do not separate world from task}

We ask whether the shared post-attention geometry is architecture-specific, training $60$ recurrent baselines on the same data (LSTM/GRU cells, $L\!\in\!\{1,2\}$ layers, $H\!\in\!\{16,\ldots,256\}$; Appendix~\ref{app:rnn_comparison}). The RNNs reach the same Bayes-optimal loss, and they too must track the world-state: by construction, Walker~O's positions $(0,0)$ and $(\pm2,0)$ share a next-step distribution but diverge later, so an optimal predictor has to keep them apart across the recurrence, beyond what $\vq_t$ alone encodes. Linear probes confirm that both $\vz_t=[x_t/t,\,y_t/t,\,\log t]$ and $\vq_t$ are recoverable from the hidden states. What the RNNs do not reproduce is the transformer's \emph{separation}: the geometric coordinates span the transformer's post-attention residual almost fully but only partially the RNN hidden states (Appendix~\ref{app:rnn_comparison}), which carry the statistic mixed with other variance rather than isolating it as a clean, variance-dominant stage. The world/task split is therefore architecture-specific, even though both architectures must compute the same world-state: attention makes a separate, explicit world-stage cheap. The RNN family is itself internally homogeneous across cell type, depth, and width (Figure~\ref{fig:cka_residual_all_walkers}), a weaker within-architecture universality \cite{maheswaranathanUniversalityIndividuality2019}.

\subsection{Predictive geometry as a world model}
The post-attention coordinates $[x_t/t,\,y_t/t,\,\log t]$ (Sections~\ref{sec:what_is_encoded}--\ref{sec:universality}) are at once a sufficient statistic for prediction and a world-state representation of the grid—world coordinates that compactly parameterize the entire future distribution. The grid walkers thus give an analytically tractable instance of the emergent world models found in sequence models trained on games and spatial or graph-structured data \cite{Li2022EmergentWorldRepresentationsOthello, Gurnee2023LanguageModelsRepresentSpaceTime, Ivanitskiy2023StructuredWorldRepresentationsMazes, Karvonen2024EmergentWorldModelsChess}, and are especially close to map-like representations learned from random walks \cite{baumgartnerCognitiveMaps2025, park2025iclr}—except that here the map is an explicit predictive statistic. This connects to work casting autoregressive models as encoders of belief states \cite{shaiTransformersRepresentBelief2024, piotrowskiConstrainedBeliefUpdates2025, riechersNeuralNetworksLeverage2025} or sufficient statistics \cite{zhangWhatShouldEmbeddings2025} of the latent process, with the twist that ours is a spatial coordinate system.

The model thus tracks world-state beyond what the next-step loss requires: for Walker~O distinct positions can share a next-step distribution yet are separated by the spatial coordinates (Section~\ref{sec:what_is_encoded}); for Walker~L the origin $(0,0)$ is reachable at different times, sharing a next-step distribution, yet is separated through $\log t$. Crucially, attention performs this state update exactly despite treating the sequence in parallel: uniform causal averaging implements the closed-form world-state transition rather than storing a static feature.

That similarly sized RNNs solve the task without adopting these coordinates makes the representation architecture-dependent—a Rashomon effect, in which equally predictive models organize information differently \cite{breimanStatisticalModelingTwo2001, rudinInterpretableMachineLearning2021, fisherAllModelsAre2019, damourUnderspecificationPresents2022}: 
the objective fixes the optimal next-step distribution but not the representation that computes it, 
and self-attention simply selects a basis in which the world-state update is cheap.

\section{Limitations and future challenges}
\label{sec:limitations}
Our toy system is deliberately simple, which makes the analytical examination above possible but is also the main limitation: scaling to settings where the predictive structure has no closed form will require a method to decompose complex distributions into tractable parts (see the analysis of \cite{shai2026transformerslearnfactoredrepresentations} for an example where this can be possible).

Generation and prediction are also confounded here: because state transitions are deterministic given a walk step, the optimal predictor coincides with the most parsimonious generator. Settings with fundamentally hidden states \cite{shaiTransformersRepresentBelief2024} or richer formal languages such as CFGs \cite{deletangNeuralNetworksChomsky2023, stroblWhatFormalLanguages2024} would test whether the geometric story persists once this confound is removed.

Finally, we do not test generalization under any distribution shift. The simple attention algorithm makes length generalization plausible for at least part of the computation \cite{zhouWhatAlgorithmsCan2023}, though extrapolation is not automatic \cite{hahnSensitiveFunctions2024}; directly probing it and other generalization regimes is a natural next step.

\section{Conclusions}
\label{sec:conclusions}
Across three walker variants, trained transformers reach Bayes-optimal next-step loss while representing more than that loss requires. The grid walker separates a world from the tasks defined on it: the spatiotemporal state $(t,\vx_t)$ is sufficient for the full future of every variant, whereas the next-step objective depends only on the less informative effective field $\vq_t$. We find that the network builds the full world-state regardless, exposing it most cleanly just after attention as a coordinate system $[x_t/t,\, y_t/t,\, \log t]$ that is shared across variants (CKA $0.95$–$0.97$), and only afterward does the MLP fold this shared geometry into each variant's next-step distribution. Attention does not merely store this state but operates on it: uniform causal averaging computes the normalized displacement, and with it the world's state transition in closed form.

Two readings of the post-attention representation therefore coincide: it is at once a world-state and a sufficient statistic for prediction, so that in learning the stochastic prediction task, the network learns the deterministic geometry beneath it. That this representation is richer than the next step requires is not a failure of compression, since it already collapses the entire history of paths into a few coordinates; the surplus world-state detail it keeps instead reflects architectural bias, as attention recovers that state cheaply whatever the active constraint. RNNs confirm that the architectural part is the geometry, not the world-state: they reach the same loss and must track the same world-state, but never isolate it as a separate, explicit coordinate stage. Whether a predictor renders the world it tracks as a distinct geometry is therefore an architectural choice, not a consequence of optimal prediction.

{
\small
\bibliographystyle{unsrtnat}
\bibliography{bibliography}
}

\appendix
\section{Computation of Probability Distributions}
\label{app:distrib_comp}

All probability distributions used for training and validation are computed exactly via a dynamic-programming recursion over the path-count propagator, with no sampling. We follow the notation of Section~\ref{sec:grid_theory}: walkers are parameterized by an endpoint set $\mathcal{P}\subset\mathbb{Z}^2$ and obstacle set $\mathcal{O}\subset\mathbb{Z}^2$ (possibly empty), with $\tau = T - t$ the remaining number of steps after a length-$t$ prefix.

\paragraph{Path-count propagator.}
For an endpoint $\vp \in \mathcal{P}$, let
\begin{equation*}
  H_\mathcal{O}^{\vp}(\tau, z) \;:=\; \#\{\text{length-}\tau\ \text{paths from}\ z\ \text{to}\ \vp\ \text{avoiding}\ \mathcal{O}\}.
\end{equation*}
This is the explicit-endpoint form of the main-text propagator $H_\mathcal{O}(\tau, z)$ of Section~\ref{sec:grid_theory}. It satisfies the linear four-neighbor recursion
\begin{equation}
  H_\mathcal{O}^{\vp}(\tau+1, z) \;=\; \mathbf{1}_{z \notin \mathcal{O}} \sum_{\Delta \in \{\Delta_L, \Delta_R, \Delta_D, \Delta_U\}} H_\mathcal{O}^{\vp}(\tau, z + \Delta),
  \label{eq:greens_recursion}
\end{equation}
with terminal condition $H_\mathcal{O}^{\vp}(0, z) = \mathbf{1}_{z = \vp}$. In the obstacle-free case ($\mathcal{O} = \emptyset$), the diagonal coordinates $u_z := x_z + y_z$, $v_z := x_z - y_z$ (and $p_u := p_x + p_y$, $p_v := p_x - p_y$ for the endpoint; cf.\ Section~\ref{sec:grid_theory}) decouple the recursion and the propagator factorizes in closed form,
\begin{equation*}
  H_\emptyset^{\vp}(\tau, z) \;=\; \binom{\tau}{(\tau + p_u - u_z)/2}\,\binom{\tau}{(\tau + p_v - v_z)/2},
\end{equation*}
nonzero only when both binomial arguments are integers in $[0, \tau]$. With obstacles no closed form exists; the recursion is iterated forward from $\tau = 0$ to $\tau = T$ and cached at every level.

\paragraph{Prefix distribution.}
Conditioned on the endpoint $\vp$, all valid length-$T$ paths from origin to $\vp$ avoiding $\mathcal{O}$ are equiprobable, so the marginal probability of a length-$t$ prefix $s_t$ ending at position $\vx_t$ is the fraction of full paths that begin with $s_t$:
\begin{equation}
  \Pr(s_t \mid \vp) \;=\; \frac{H_\mathcal{O}^{\vp}(\tau, \vx_t)}{H_\mathcal{O}^{\vp}(T, \mathbf{0})}.
  \label{eq:seq_marginal}
\end{equation}
This is proportional to the number of valid completions from $\vx_t$ to $\vp$, and is therefore not uniform in $s_t$.

\paragraph{Endpoint posterior.}
For multi-endpoint walkers, the posterior over endpoints given the prefix follows from Bayes' rule with a uniform endpoint prior,
\begin{equation*}
  w_t(\vp \mid \vx_t) \;=\; \frac{\Pr(s_t \mid \vp)}{\sum_{\vp' \in \mathcal{P}} \Pr(s_t \mid \vp')},
\end{equation*}
and the unconditional prefix marginal is $\Pr(s_t) = |\mathcal{P}|^{-1} \sum_{\vp} \Pr(s_t \mid \vp)$.

\paragraph{Next-step probabilities.}
By Equation~\ref{eq:greens_recursion}, the conditional next-step law given endpoint $\vp$ is the finite-horizon Doob $h$-transform, with $H_\mathcal{O}^{\vp}$ playing the space-time harmonic function \cite{doobConditionalBrownianMotion1957,levinMarkovChainsMixing2017}:
\begin{equation}
  \Pr(\Delta \mid s_t; \vp) \;=\; \frac{H_\mathcal{O}^{\vp}(\tau - 1, \vx_t + \Delta)}{H_\mathcal{O}^{\vp}(\tau, \vx_t)},
  \label{eq:next_step_exact}
\end{equation}
which automatically vanishes when $\vx_t + \Delta \in \mathcal{O}$ and is well-defined because the four numerator terms sum to the denominator. The multi-endpoint distribution is the posterior mixture $\Pr(\Delta \mid s_t) = \sum_{\vp} w_t(\vp \mid \vx_t)\,\Pr(\Delta \mid s_t; \vp)$.

\paragraph{Implementation.}
The recursion is run in diagonal coordinates, where the four-neighbor update becomes four corner additions on a $(T+1) \times (T+1)$ array and the parity constraint $u_z + v_z \equiv \tau \pmod 2$ is enforced automatically. All arithmetic is performed in log-space. 
Next-step tables for every equivalence class $(\vx_t, t)$ are computed once via Equation~\ref{eq:next_step_exact} and cached as training batches and validation targets.

\section{Closed-Form Expressions for the Effective Field}
\label{app:gaussian_counts}
\label{app:suff_vec}
\label{app:eff_field}

This appendix supplies the two closed-form results referenced from Section~\ref{sec:grid_theory}: explicit $\mathrm{artanh}$ expressions for the Walker~L effective field, and the bridge-field interpretation of the next-step moments for Walker~D.

\paragraph{Walker~L: closed-form artanh.}
For a single endpoint $\vp$ and $\mathcal{O}=\emptyset$, the next-step distribution factorizes as in Equation~\ref{eq:next_step_probs_product}:
\begin{equation*}
  \Pr(\Delta\mid s_t;\vp) \;=\; \tfrac{1}{4}\bigl(1+\mu_p\,\Delta u\bigr)\bigl(1+\nu_p\,\Delta v\bigr),
  \qquad \mu_p = \frac{p_u - u_t}{\tau},\quad \nu_p = \frac{p_v - v_t}{\tau}.
\end{equation*}
Taking logarithms and applying the identity $\log(1\pm z) = \tfrac{1}{2}\log(1-z^2)\pm \mathrm{artanh}(z)$ separately to the two factors,
\begin{equation*}
  \log\Pr(\Delta\mid s_t;\vp) \;=\; \mathrm{artanh}(\mu_p)\,\Delta u + \mathrm{artanh}(\nu_p)\,\Delta v + C(\mu_p,\nu_p),
\end{equation*}
with offset $C(\mu_p,\nu_p) = \tfrac{1}{2}\log[(1-\mu_p^2)(1-\nu_p^2)] - \log 4$ independent of $\Delta$. Matching against Equation~\ref{eq:eff_field} reads off the three components of the effective field,
\begin{equation}
  q_u \;=\; -\mathrm{artanh}(\mu_p),\qquad
  q_v \;=\; -\mathrm{artanh}(\nu_p),\qquad
  q_{uv} \;=\; 0,
  \label{eq:walker_l_qfield}
\end{equation}
confirming that Walker~L's effective field is effectively 2D. The $\mathrm{artanh}$ logits diverge at the admissible-domain boundary $|\mu_p|=1$ or $|\nu_p|=1$, where one or more next steps become impossible and the Softmax form of Equation~\ref{eq:next_step_probs} is the appropriate limit.

\paragraph{Walker~D: bridge-field interpretation.}
For two endpoints $\mathcal{P} = \{\vp_+,\vp_-\}$ with posterior weights $w_\pm = w_t(\vp_\pm\mid\vx_t)$ ($w_+ + w_- = 1$), the next-step distribution is the posterior mixture (Equation~\ref{eq:multi_endpoint_mix}). Linearity of expectation through the product form of Equation~\ref{eq:next_step_probs_product} immediately gives the next-step moments as posterior averages of the per-endpoint bridge fields,
\begin{equation*}
  \langle\Delta u\rangle = \langle\mu\rangle_\mathcal{P},\qquad
  \langle\Delta v\rangle = \langle\nu\rangle_\mathcal{P},\qquad
  \langle\Delta u\,\Delta v\rangle = \langle\mu\nu\rangle_\mathcal{P},
\end{equation*}
where $\langle f\rangle_\mathcal{P} := \sum_{\vp} w_t(\vp\mid\vx_t)\,f(\vp)$ and $\mu, \nu$ are the per-endpoint drifts of Equation~\ref{eq:next_step_probs_product}. Writing $s := w_+ - w_- \in [-1, 1]$ for the belief asymmetry, the third moment satisfies the covariance identity
\begin{equation}
  \langle\mu\nu\rangle_\mathcal{P} - \langle\mu\rangle_\mathcal{P}\langle\nu\rangle_\mathcal{P}
  \;=\; \tfrac{1}{4}(1-s^2)\,(\mu_+ - \mu_-)(\nu_+ - \nu_-),
  \label{eq:walker_d_cov}
\end{equation}
which is nonzero whenever both endpoints retain posterior support ($|s|<1$) and their bridge drifts differ in both components. For the symmetric Walker~D configuration $\vp_\pm = (\pm 4, 0)$ used in our experiments, $\mu_+ - \mu_- = \nu_+ - \nu_- = 8/\tau$ and Equation~\ref{eq:walker_d_cov} specializes to $16(1-s^2)/\tau^2$. The product factorization of Equation~\ref{eq:next_step_probs_product} is therefore broken at every prefix where the endpoint posterior is non-degenerate, so $q_{uv}\neq 0$ in the effective-field decomposition (Equation~\ref{eq:eff_field}) and all three components of $\vq_t$ carry independent information about position and belief.

\section{First-Layer Attention Mechanism}
\label{app:attention_mechanism}

We derive the coordinates produced by the first attention layer (Section~\ref{sec:first_attention_mechanism}). Following the transformer-circuits decomposition of attention heads into attention-routing and value-writing components \cite{elhageMathematicalFramework2021}, for a source position $i$ and token $c_i$, the value-write vector is $w(c_i,i) := v(c_i,i) W_O$, where $v(c_i,i)$ is the head's value vector and $W_O$ is the output matrix. Although the value vector is computed after LayerNorm and could in principle couple token and position nonlinearly, empirically we observe an additive main-effects decomposition:
\begin{equation}
  w(c_i,i) \;\approx\; b + a_{c_i} + p_i ,
  \label{eq:value_main_effects}
\end{equation}
where $a_{c_i}$ is a token code and $p_i$ is a position code. The token--position interaction accounts for less than $2\times 10^{-5}$ of the centered value-write variance in the Walker~L model shown in Figure~\ref{fig:first_layer_encoding}.

The attention pattern is nearly uniform over the causal prefix: at query position $t$, the head approximately averages values from positions $1,\ldots,t$. Combining uniform averaging with Equation~\ref{eq:value_main_effects} gives
\begin{equation}
  \mathbf{o}^{\mathrm{attn}}_t
  \;\approx\;
  b
  + \frac{1}{t}\sum_{i=1}^t a_{c_i}
  + \frac{1}{t}\sum_{i=1}^t p_i .
  \label{eq:attention_average_decomposition}
\end{equation}
The token-average term equals $A\,\vx_t/t$ for some linear map $A$ (Equation~\ref{eq:compass_avg}). The position-average term,
\begin{equation}
  h_t := \frac{1}{t}\sum_{i=1}^t p_i ,
\end{equation}
is path-independent and depends only on $t$. Figure~\ref{fig:first_layer_encoding}C verifies the decomposition: after removing the best-fit affine $x_t/t,y_t/t$ component from the empirical attention output, the residual time centroids coincide with $h_t$.

\section{Representational Analysis}
\label{app:rep_analysis}

\subsection{Metrics}
\label{app:metrics}
We characterize the alignment between an activation $\va_\ell(s_t)\in\mathbb{R}^{d_\ell}$ at hook $\ell$ and a target vector $\vy(s_t)\in\mathbb{R}^d$ (typically either the geometric feature vector $\vz_t=[x_t/t,\,y_t/t,\,\log t]$ or the effective field $\vq_t$, see Appendix~\ref{app:eff_field}) using two complementary tools. Both restrict to linear structure in the activations, in line with the empirical observation that task-relevant features in trained networks tend to be encoded as linear directions on which downstream components act \cite{park2023linear}.

\paragraph{Affine probe and $R^2$.}
We fit an affine map $\va \mapsto W\va + \vb$ (or its converse $\vy \mapsto W'\vy + \vb'$) by ordinary least squares (OLS) and report the joint $R^2$ across all output dimensions. The two probe directions are not redundant and answer different questions:
\begin{itemize}
  \item \textbf{Decoding} ($\va \to \vy$). How recoverable is the target from a linear function of the activation? A high decoding $R^2$ means $\vy$ lies (up to noise) in the column space of $\va$; it does not constrain what else $\va$ may contain.
  \item \textbf{Encoding} ($\vy \to \va$). How much of the activation variance is contained in the affine image of the target? Provided $d \ll d_\ell$, a high encoding $R^2$ constrains $\va$ to mostly lie in a low-dimensional affine subspace spanned by $\vy$.
\end{itemize}
Both directions appear in this paper; the relevant text and figure captions state which direction (and which aggregation and which evaluation, see below) is plotted in each panel.

\paragraph{Aggregation: state means vs.\ raw datapoints.}
The dataset of prefixes naturally partitions into equivalence classes indexed by $(\vx_t, t)$: two prefixes that reach the same lattice position at the same time give rise to the same predictive distribution and---since $\vz_t$ and $\vq_t$ depend only on $(\vx_t,t)$---to the same target. The two probe directions motivate different aggregation choices.

For the \textbf{encoding} direction (target $\to$ activations) we use state-mean activations: one averaged activation per equivalence class. Different paths to the same $(\vx_t,t)$ produce the same feature vector but may produce slightly different activations (since the model reads the full sequence). Fitting on raw datapoints would therefore conflate the between-state structure we care about with within-state activation variation that is, by construction, unexplainable by the features. State means remove this noise and give each state equal weight, making the encoding $R^2$ a clean test of whether a low-dimensional feature set spans the activation geometry.

For the \textbf{decoding} direction (activations $\to$ target) we use raw per-position datapoints, to be faithful to the distribution the network was actually trained on. The network processes individual sequences, not state averages, so asking whether the target is recoverable from a single-path activation is the natural question. Raw decoding has $n_{\rm samples} \gg d_{\rm source}$ at every hook and width considered, so OLS is well-conditioned without regularization. Figure~\ref{fig:r2_combined}A, Figure~\ref{fig:activation_to_feature_r2_all_dheads}, and Figure~\ref{fig:rnn_capacity_r2} use this aggregation; all other figures use state means.

\paragraph{In-sample vs.\ held-out evaluation.}
For each probe we report either an in-sample $R^2$ on the full set or a held-out estimate from $10$-fold cross-validation (folds are random partitions of the input rows of each probe---state classes for encoding fits, raw datapoints for decoding fits). The held-out value is the unbiased measure of alignment and is what we use whenever the comparison across architectures or variants matters: Figure~\ref{fig:r2_combined}C, Figure~\ref{fig:features_grid}, Table~\ref{tab:feature_r2_arch}, and Figure~\ref{fig:rnn_capacity_r2}. The in-sample $R^2$ suffices for diagnostic plots that characterize a single representation per hook: Figure~\ref{fig:r2_combined}A,B and Figure~\ref{fig:activation_to_feature_r2_all_dheads}. Captions state which is reported.

\paragraph{Linear centered kernel alignment (lCKA).}
We measure run-to-run representational similarity with linear CKA \cite{kornblithSimilarityNeuralNetwork2019}, computed on (mean-centered) state-mean activations of two runs $r, r'$:
\begin{equation*}
    \mathrm{lCKA}\bigl(\bar{A}_r, \bar{A}_{r'}\bigr) = \frac{\|\bar{A}_r^{\top} \bar{A}_{r'}\|_F^2}{\|\bar{A}_r^{\top} \bar{A}_r\|_F\,\|\bar{A}_{r'}^{\top} \bar{A}_{r'}\|_F},
\end{equation*}
with $\bar{A}_r \in\mathbb{R}^{n_{\rm states}\times d_r}$. All lCKA values reported in the paper are computed on state-mean activations.
For cross-variant transformer comparisons, all runs are first restricted to the same global support intersection of valid prefixes, and rows are then averaged by shared $(\vx_t,t)$ state class. Thus each CKA value compares two representations of the same ordered set of state classes.

\paragraph{$R^2$ vs.\ lCKA.}
The key difference between the two metrics is that $R^2$ ignores the variance structure of the source space (it maps from sources to targets regardless of which directions carry variance), whereas lCKA down-weights directions of low variance in the source space, thus erasing the distinction between source and target.

\subsection{Additional diagnostics}
\label{app:additional_diagnostics}

\begin{figure}[t]
  \centering
  \includegraphics[width=\linewidth]{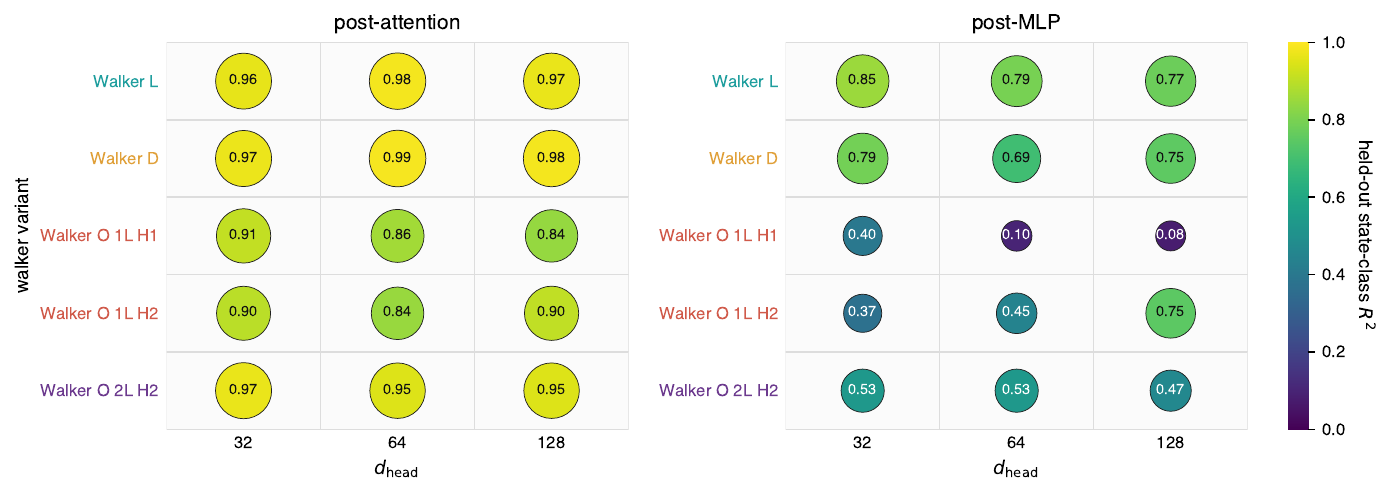}
  \caption{Held-out state-class $R^2$ of a linear probe onto the 3D geometric features $[x_t/t, y_t/t, \log t]$, by selected walker/architecture row and head dimension (columns). Left: post-attention residual stream. Right: post-MLP residual stream. Table~\ref{tab:feature_r2_arch} reports the complete architecture sweep.}
  \label{fig:features_grid}
\end{figure}

\paragraph{Held-out feature recovery across $d_{\rm head}$.}
Figure~\ref{fig:features_grid} extends panel~C of Figure~\ref{fig:r2_combined} across $d_{\rm head}$ for a compact set of architecture selections. The qualitative pattern is the same at every width: post-attention $R^2$ stays high across walker variants, and post-MLP $R^2$ drops, with the strongest drop on the most constrained Walker~O-1L H1 model. Table~\ref{tab:feature_r2_arch} reports the complete $L\in\{1,2\}$, $H\in\{1,2\}$, $d_{\rm head}\in\{32,64,128\}$ sweep for all three walkers; across all $36$ models, post-attention $R^2$ lies in the $0.84$--$0.99$ range.

\begin{table}[t]
  \centering
  \small
  \setlength{\tabcolsep}{4.2pt}
  \begin{tabular}{lcc|cc|cc|cc}
    \toprule
    & & & \multicolumn{2}{c|}{$d_{\rm head}=32$} & \multicolumn{2}{c|}{$d_{\rm head}=64$} & \multicolumn{2}{c}{$d_{\rm head}=128$} \\
    walker & $L$ & $H$ & attn & MLP & attn & MLP & attn & MLP \\
    \midrule
    L & 1 & 1 & 0.962 & 0.850 & 0.982 & 0.793 & 0.967 & 0.775 \\
    L & 1 & 2 & 0.985 & 0.733 & 0.969 & 0.755 & 0.937 & 0.707 \\
    L & 2 & 1 & 0.955 & 0.856 & 0.988 & 0.825 & 0.952 & 0.811 \\
    L & 2 & 2 & 0.988 & 0.791 & 0.951 & 0.785 & 0.934 & 0.785 \\
    D & 1 & 1 & 0.969 & 0.785 & 0.988 & 0.695 & 0.981 & 0.750 \\
    D & 1 & 2 & 0.986 & 0.727 & 0.979 & 0.757 & 0.972 & 0.797 \\
    D & 2 & 1 & 0.965 & 0.805 & 0.991 & 0.770 & 0.974 & 0.662 \\
    D & 2 & 2 & 0.990 & 0.700 & 0.977 & 0.644 & 0.962 & 0.684 \\
    O & 1 & 1 & 0.907 & 0.402 & 0.861 & 0.101 & 0.842 & 0.078 \\
    O & 1 & 2 & 0.897 & 0.368 & 0.845 & 0.448 & 0.903 & 0.749 \\
    O & 2 & 1 & 0.851 & 0.722 & 0.936 & 0.474 & 0.925 & 0.377 \\
    O & 2 & 2 & 0.967 & 0.534 & 0.946 & 0.534 & 0.947 & 0.468 \\
    \bottomrule
  \end{tabular}
  \caption{Complete architecture-resolved held-out state-class $R^2$ for linear recovery of the geometric features $[x_t/t, y_t/t, \log t]$ across the $36$ trained transformer models. ``attn'' denotes the post-attention residual stream and ``MLP'' denotes the post-MLP residual stream. The consistently high post-attention values indicate that the attention block makes this sufficient statistic linearly available across architectures. The lower post-MLP values show that the MLP has already applied a variant-specific nonlinear transformation, so the activations are no longer as faithfully linearly obtainable from the coordinates. Figure~\ref{fig:features_grid} gives a compact visual summary of this diagnostic, while this table makes the layer count $L$ and number of attention heads $H$ explicit for every architecture in the sweep.}
  \label{tab:feature_r2_arch}
\end{table}

\begin{figure}[t]
  \centering
  \includegraphics[width=\linewidth]{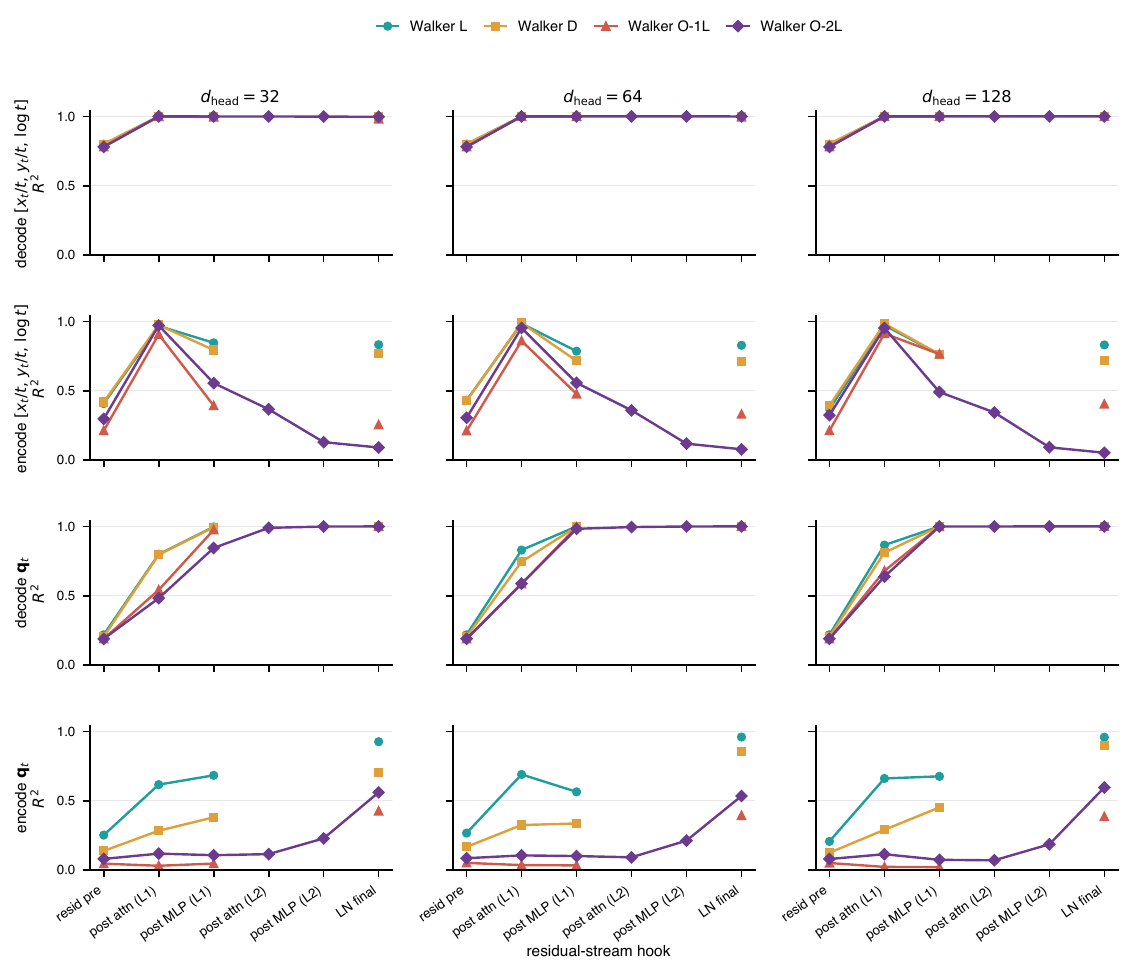}
  \caption{All-walk in-sample $R^2$ for affine probes between residual-stream activations and target features, across $d_{\rm head}\in\{32,64,128\}$. At each width we show the four main-text selected runs (Walker L L1H1, Walker D L1H1, Walker O L1H2, and Walker O L2H2), not every network in the architecture sweep. Decode rows fit the geometric target $\vz_t=[x_t/t,y_t/t,\log t]$ or effective field $\vq_t$ from raw per-position activations; encode rows fit the state-mean activation from the corresponding state-mean target. Gaps mark second-layer hooks, absent in one-layer runs.}
  \label{fig:activation_to_feature_r2_all_dheads}
\end{figure}

\paragraph{Feature decodability across $d_{\rm head}$.}
Figure~\ref{fig:activation_to_feature_r2_all_dheads} reports in-sample affine probes in both directions between residual-stream activations and the two target feature sets. Decode rows fit targets from raw prefix-position activations, while encode rows fit state-mean activations from state-mean targets, as in Figure~\ref{fig:rnn_capacity_r2}. The panel uses the same main-text selected runs at each width (Walker L and Walker D with one layer and one head, Walker O with one layer and two heads, and Walker O with two layers and two heads), rather than the full architecture sweep. The geometric features are linearly decodable after attention across widths, and the effective field $\vq_t$ is nearly perfectly decodable after the MLP and final LayerNorm; the encode rows show how much of the activation variance is spanned by each target.

\begin{figure}[t]
  \centering
  \includegraphics[width=\linewidth]{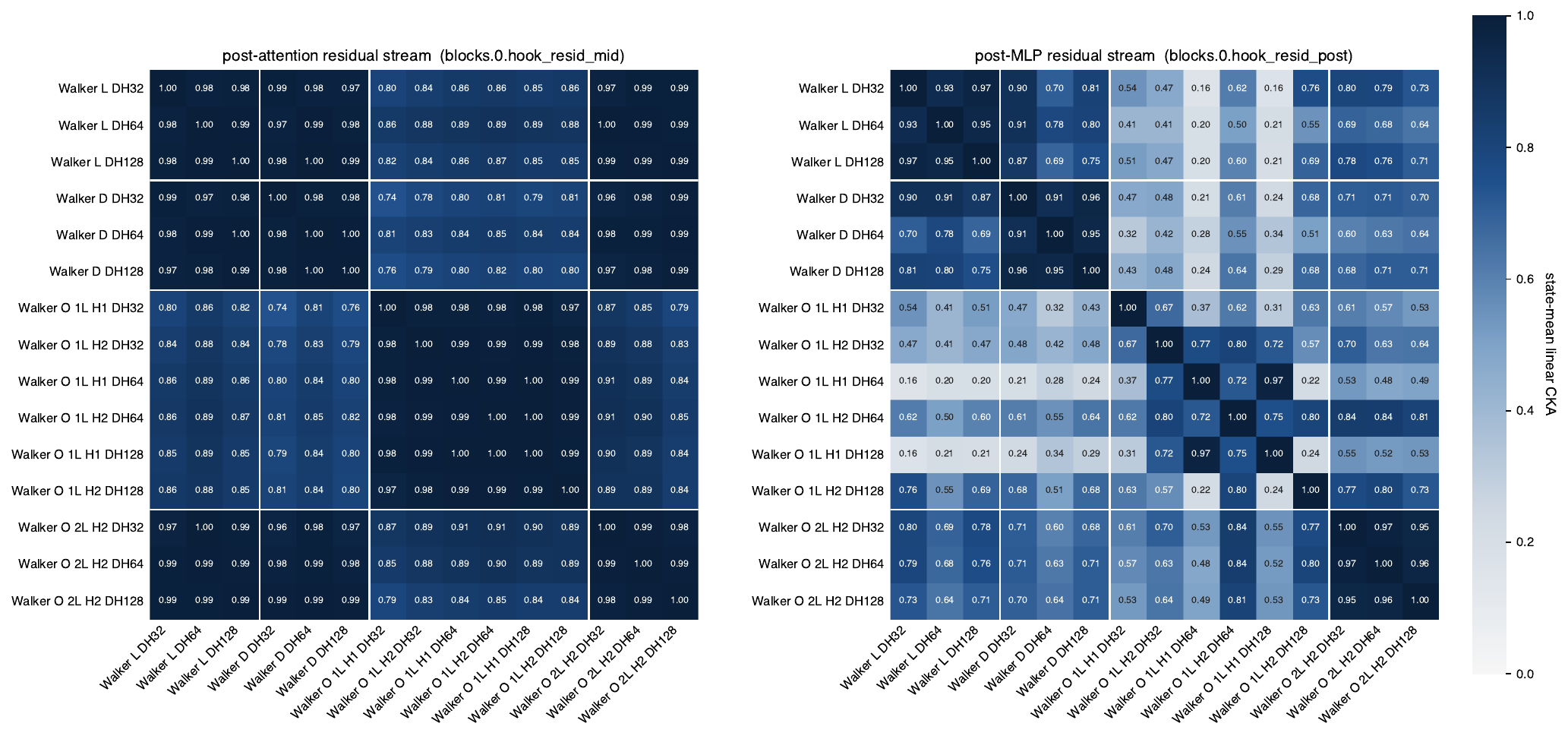}
  \caption{Full pairwise linear CKA matrix between state-mean activations of every individual run in our sweep, at the post-attention residual stream (left) and the post-MLP residual stream (right).}
  \label{fig:cka_full}
\end{figure}

\paragraph{Pairwise CKA across all individual runs.}
Figure~\ref{fig:cka_full} reports the full pairwise linear CKA matrix between state-mean activations of every trained network in our sweep, without grouping by walker variant. Each row and column is one run, labeled by walker and architecture. The block structure of the matrix reflects the universality result of Section~\ref{sec:universality} at run level: at the post-attention residual stream most pairs are highly aligned, while at the post-MLP residual stream the cross-variant blocks lose alignment.

\begin{figure}[t]
  \centering
  \includegraphics[width=\linewidth]{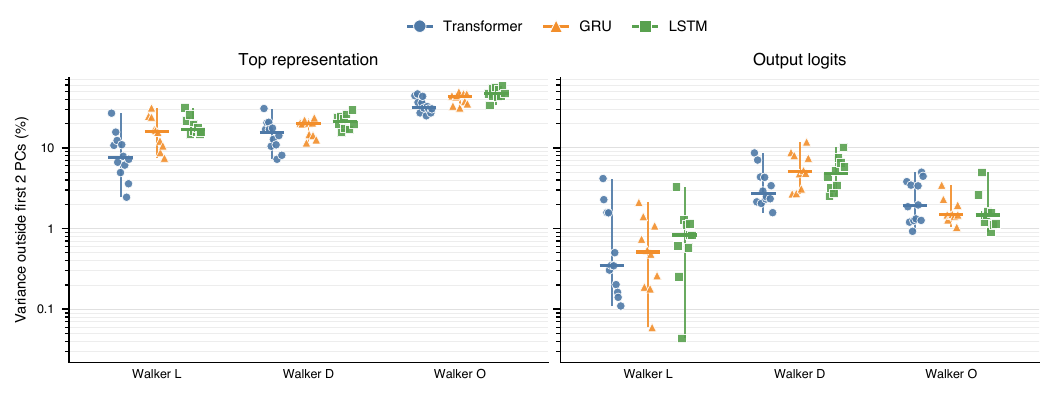}
  \caption{Variance outside the best two-dimensional PCA subspace for the top representation (left: transformer final LayerNorm activations or RNN top hidden states) and raw output logits (right). Each point is one final checkpoint. Color and marker shape denote architecture family; vertical bars show the within-family min--max range for each walker, and horizontal ticks show the median.}
  \label{fig:output_logit_pca_dim}
\end{figure}

\paragraph{Top-representation and output-logit PCA dimension.}
The geometric pipeline in Figure~\ref{fig:predictive_transform} suggests that the final readout is close to two-dimensional. To compare this low-dimensionality before and after the readout, Figure~\ref{fig:output_logit_pca_dim} measures the variance outside the best two-dimensional PCA subspace for the top representation (transformer final LayerNorm activations, or the top recurrent hidden state for RNNs) and for the raw output logits across the transformer and GRU/LSTM sweeps.

\subsection{RNN comparison}
\label{app:rnn_comparison}
This appendix gathers the diagnostic plots underlying the cross-architecture comparison summarized at the end of Section~\ref{sec:universality}. We train $60$ RNNs covering LSTM and GRU cells, $L\in\{1,2\}$ recurrent layers, hidden sizes $H\in\{16,32,64,128,256\}$, and all three walker variants. All RNNs use the same context length, optimizer, and learning-rate schedule as the transformer sweep, and validation losses are normalized by the same per-position Bayes-optimal floor used for the transformer (Tables~\ref{tab:walkers} and~\ref{tab:rnn_losses}).

\begin{figure}[t]
  \centering
  \includegraphics[width=\linewidth]{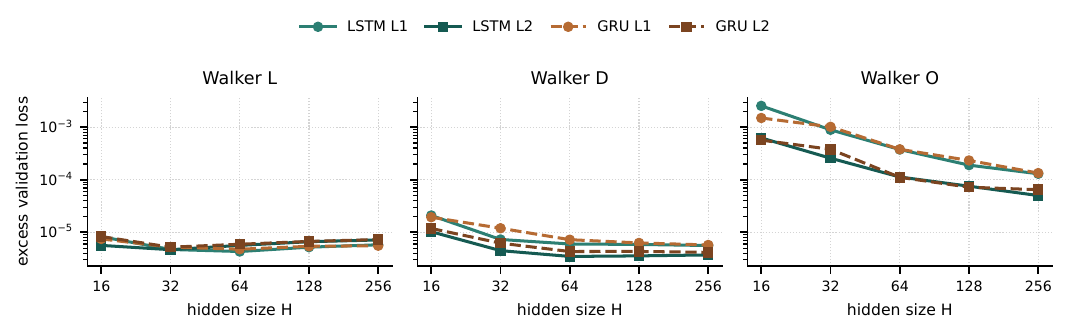}
  \caption{RNN capacity scan: predictive loss. Final validation loss is plotted as excess over the Bayes-optimal value, as a function of hidden size, recurrent cell type, depth, and walker variant. Loss is a model-level quantity, so two-layer models appear once rather than separately for each recurrent layer. All RNN families approach the optimum as width increases; Walker~O remains the hardest variant at small widths.}
  \label{fig:rnn_capacity_loss}
\end{figure}

\emph{Predictive performance.} All RNN families reach the Bayes-optimal validation loss to within a small multiple of the transformer noise floor on Walkers~L and~D; Walker~O is the only variant in which the smallest-$H$ runs leave a measurable gap (Figure~\ref{fig:rnn_capacity_loss}). The corresponding final losses and training trajectories are reported together with the transformer sweep in Section~\ref{app:training_curves}.

\begin{figure}[t]
  \centering
  \includegraphics[width=\linewidth]{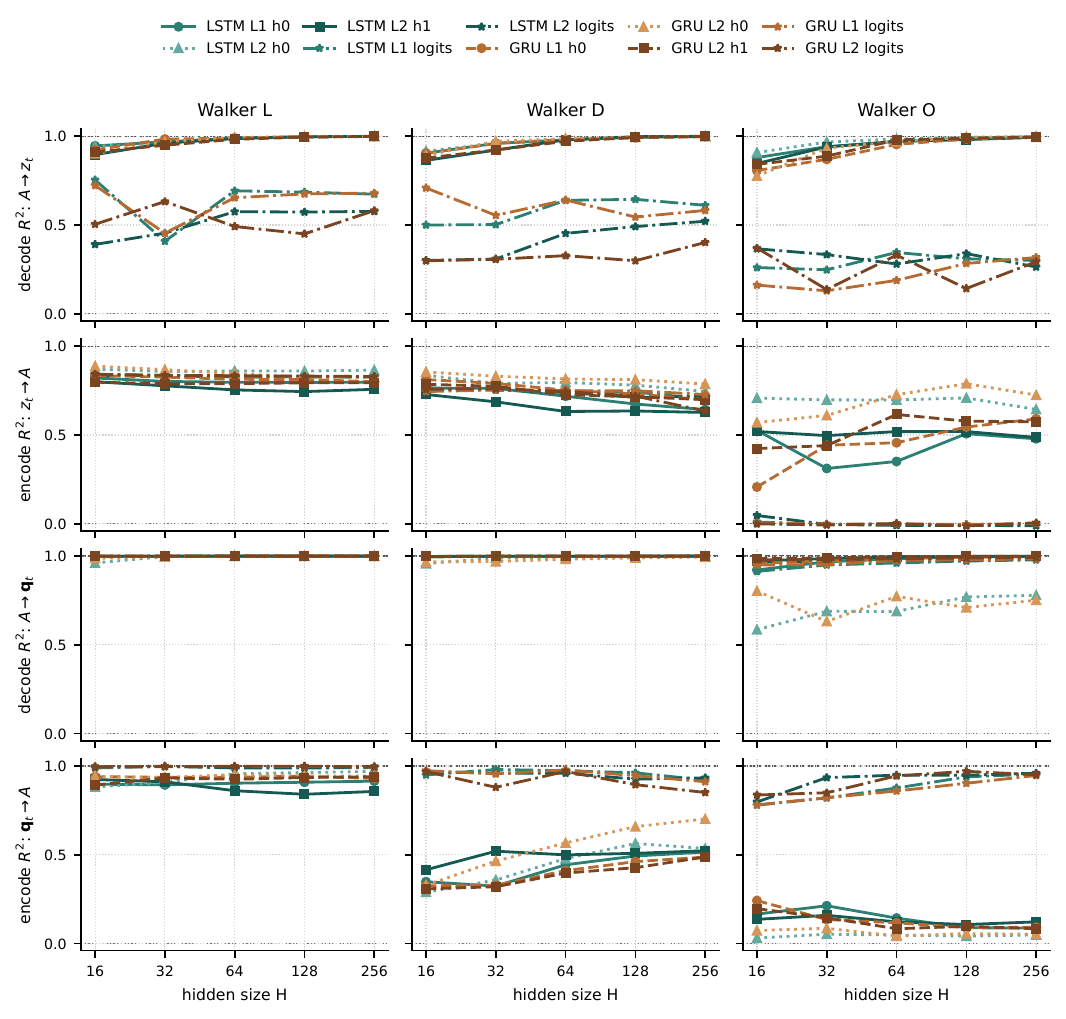}
  \caption{RNN capacity scan: linear accessibility of geometric and predictive targets. Each panel shows held-out CV $R^2$ as a function of hidden size for LSTM and GRU recurrent networks trained on the three walker variants. The target $\vz_t=[x_t/t, y_t/t, \log t]$ is the explicit geometric coordinate system used in the main text, while $\vq_t$ is the effective field. Decode rows fit the target from the raw per-position activation $A$ via OLS; encode rows fit the state-mean activation from the target via OLS. For two-layer RNNs, $h0$ and $h1$ denote the first and second recurrent hidden layers; for one-layer RNNs, $h0$ is the only hidden layer. The \emph{logits} series probes the output-layer logits as a sanity check: $\vq_t$ is a fixed linear transformation of the centered logits and saturates to $R^2=1$ at this hook for Walkers~L and~D.}
  \label{fig:rnn_capacity_r2}
\end{figure}

\begin{figure}[t]
  \centering
  \includegraphics[width=\linewidth]{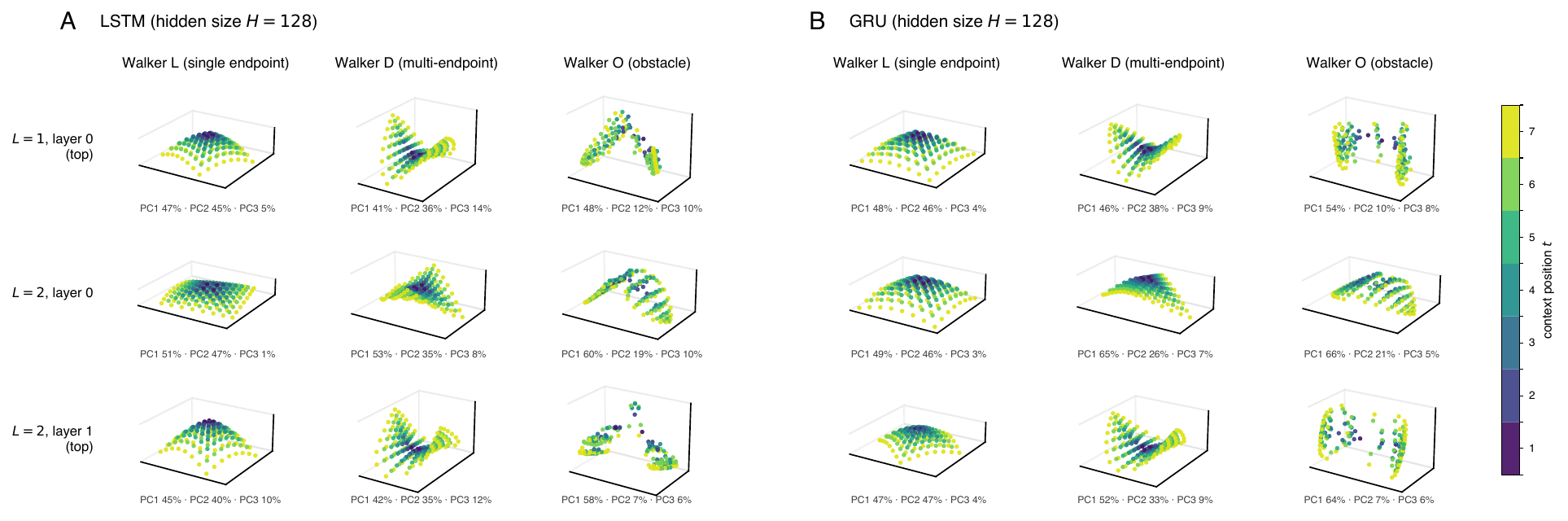}
  \caption{Hidden-state geometry across architectures, depths, and walkers at $H=128$. \textbf{(A)} LSTM. \textbf{(B)} GRU. Within each panel, rows are $L=1$ top hidden ($\equiv$ layer~$0$), $L=2$ layer~$0$, and $L=2$ layer~$1$ (top); columns are Walker~L, Walker~D, and Walker~O. Each cell shows the 3D PCA of the state-mean recurrent activation, coloured by context position $t$, with axes drawn on equal scale in data units so a near-flat manifold renders as flat. The geometry is set primarily by the task: Walker~L collapses to a near-2D dome, Walker~D forms a 3D bowl, and Walker~O is capacity-sensitive. Neither the LSTM/GRU choice nor the depth substantially alters the manifold's intrinsic shape.}
  \label{fig:rnn_geometry_variants_combined}
\end{figure}

\emph{Linear accessibility of the sufficient statistic.} We probe each recurrent hidden state in both directions: the \emph{decode} direction maps the activation $A$ to either the 3D geometric statistic $\vz_t=[x_t/t,y_t/t,\log t]$ or the 3D effective field $\vq_t$, and the \emph{encode} direction is the converse. Decode $R^2$ is essentially saturated across all three walker variants at moderate widths, with $\vq_t$ decode approaching the output-logits ceiling at the top hidden layer for every walker (Walker~O at $H\!=\!64$ already reaches $R^2 \approx 0.99$). The encode direction is more discriminating: $\vq_t$ alone affinely spans the activation for Walker~L ($R^2 \approx 0.91$ at the top hidden layer) but only $0.43$ for Walker~D and $0.13$ for Walker~O, while the geometric statistic $\vz_t$ spans more uniformly across walkers (range $0.48$--$0.79$). Recurrent hidden states therefore embed the sufficient statistic inside a strictly larger subspace whose extra variance is most pronounced for Walker~O---consistent with the diffuse hidden-state geometry shown in Figure~\ref{fig:rnn_geometry_variants_combined}. The \emph{logits} series in Figure~\ref{fig:rnn_capacity_r2} probes the output-layer logits as a sanity check: $\vq_t$ is a fixed linear transformation of the centered logits and saturates to $R^2\!=\!1$ at this hook for the top layer of all walkers.

\emph{Geometry of the hidden-state manifold.} For a representative width $H=128$, the 3D PCA of the state-mean recurrent hidden representation is a near-2D dome on Walker~L, a 3D bowl on Walker~D, and a capacity-sensitive shape on Walker~O. The pattern reproduces nearly identically between LSTM and GRU, between $L=1$ and $L=2$, and within an $L=2$ stack between layer~$0$ and layer~$1$; depth contributes a near-isometric reparametrization rather than a structural change (Figure~\ref{fig:rnn_geometry_variants_combined}).

\begin{figure}[t]
  \label{fig:rnn_cka_allvall}
  \centering
  \includegraphics[width=\linewidth]{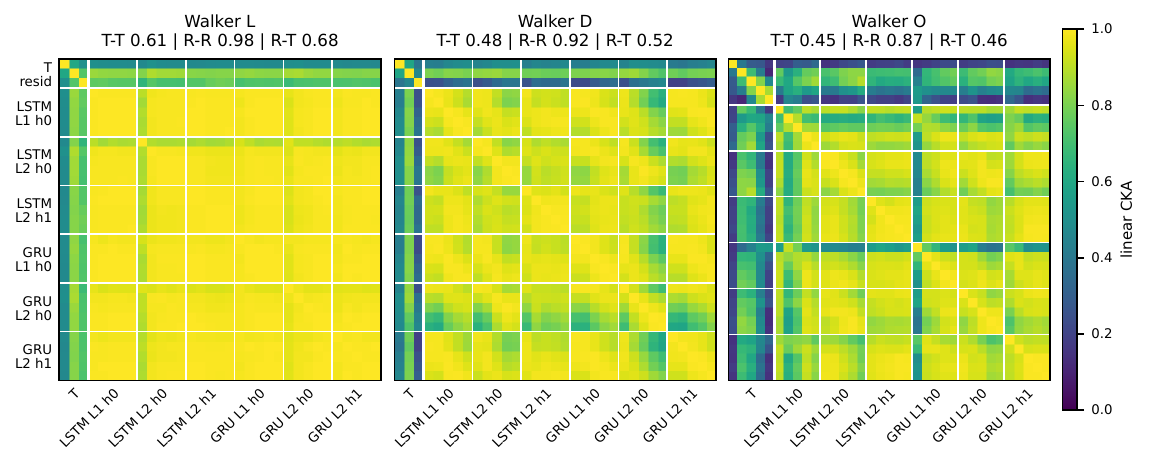}
  \caption{All-vs-all state-mean lCKA, transformer residual streams vs.\ RNN hidden layers, per walker. Rows and columns of each heatmap are: every transformer residual-stream hook of the canonical $d_{\rm head}=64$ run (the first few entries, separated by the white guide lines), followed by every per-layer hidden state of every RNN in the sweep, grouped by (cell type, depth, hidden size, layer index). Three blocks are visible per walker: the transformer's own residual stream is internally dissimilar; the RNN family is internally very similar; the off-diagonal block between the two architectures is moderate ($\approx 0.5$--$0.7$ for Walkers~L/D, lower for Walker~O). LayerNorm hooks are excluded.}
  \label{fig:cka_residual_all_walkers}
\end{figure}

\emph{Cross-architecture representational similarity.} We compute state-mean linear CKA between every transformer residual-stream hook of the canonical $d_{\rm head}=64$ run and every per-layer hidden state of every RNN in the sweep, separately for each walker (Figure~\ref{fig:cka_residual_all_walkers}). Three blocks emerge per walker. Within the transformer's own residual stream, lCKA is low---the pre-attention, post-attention, and post-MLP states have substantially different representational geometries, matching the per-stage decomposition that Section~\ref{sec:universality} resolves at the group level. Within the RNN family, lCKA is uniformly high---every recurrent hidden state across hidden sizes, depths, and the LSTM/GRU split clusters into essentially one form. The off-diagonal RNN-vs-transformer block sits at $\approx 0.5$--$0.7$ for Walkers~L and~D and lower for Walker~O. LayerNorm hooks are excluded since RNNs have no analogue.

\section{Training Details and Loss Curves}

\subsection{Hyperparameters}
\label{app:hyperparams}
All models are instances of the \texttt{HookedTransformer} architecture from the TransformerLens library \cite{nanda2022transformerlens}, decoder-only transformers with LayerNorm normalization and ReLU MLP activations. We sweep over architectures with $L\in\{1,2\}$ layers and $H\in\{1,2\}$ attention heads, with head dimension $d_{\rm head}\in\{32,64,128\}$; the model dimension is $d_{\rm model}=H\cdot d_{\rm head}$ and the MLP width is $d_{\rm mlp}=4\,d_{\rm model}$. One model is trained per configuration. The context window is fixed to $K=8$ tokens. Training ran for $10\,000$ epochs of $70$ minibatches of size $256$. Networks were trained with PyTorch's AdamW optimizer with default \texttt{betas} and \texttt{eps}—$(0.9, 0.999)$ and $10^{-8}$—and zero weight decay. The peak learning rate was $10^{-4}$, with a cosine schedule including a $3\%$ warmup that decays to $10\%$ of the peak by the end of training.

\subsection{Final losses and training curves}
\label{app:training_curves}
\begin{figure}[t]
  \centering
  \includegraphics[width=\linewidth]{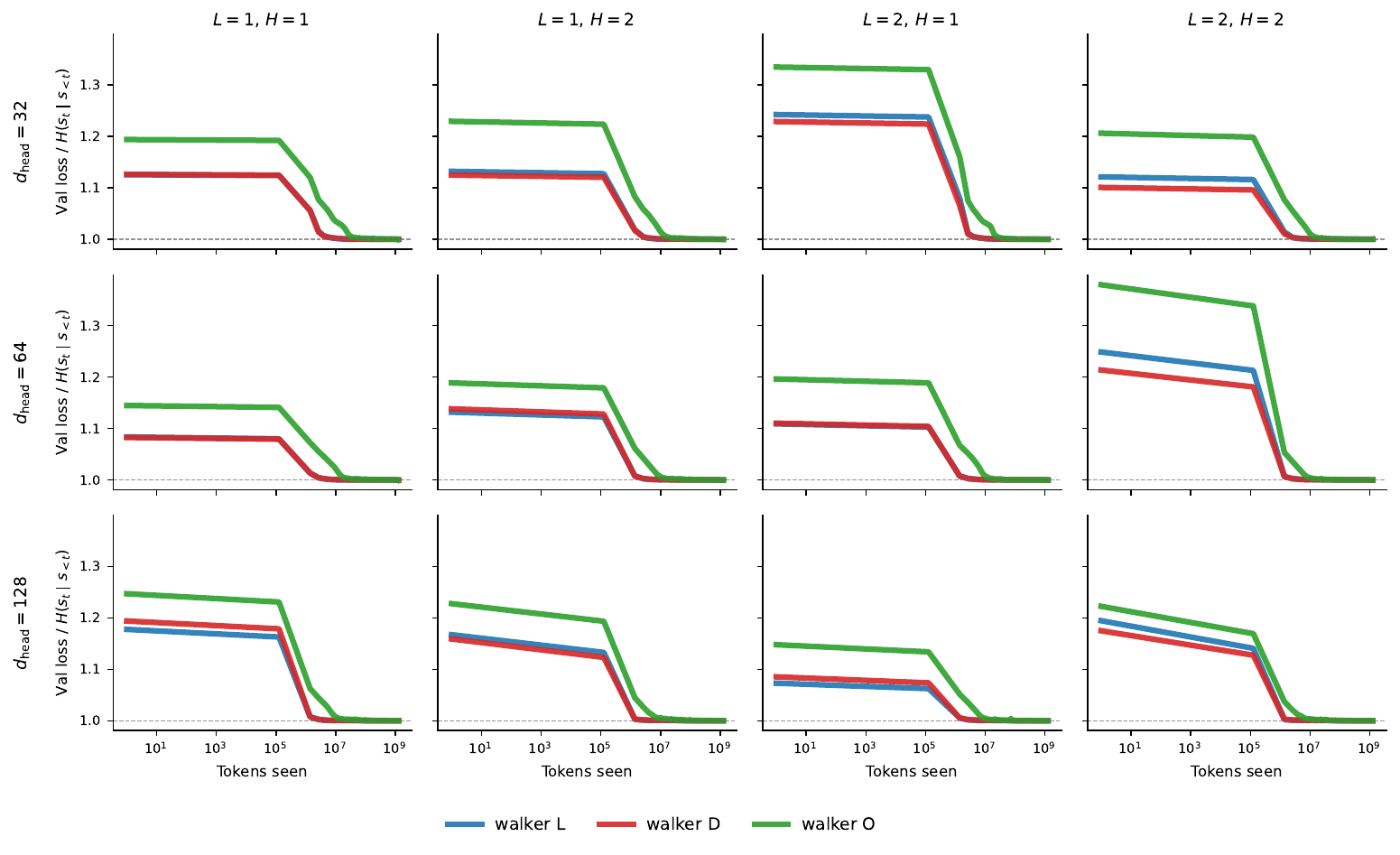}
  \caption{Validation loss vs.\ tokens seen for the full sweep. Rows are head dimensions ($d_{\rm head}\in\{32,64,128\}$); columns are layer/head configurations. Each panel shows three curves, one per walker variant. The vertical axis is the same dimensionless ratio as Table~\ref{tab:walkers}: $\mathrm{CE}_{\rm model}/H(c_{t+1}\mid s_t)$, averaged over context positions; the dashed line marks the Bayes-optimal value $1.0$. The horizontal axis is the cumulative number of training tokens (log scale).}
  \label{fig:loss_curves_grid}
\end{figure}

\begin{figure}[t]
  \centering
  \includegraphics[width=\linewidth]{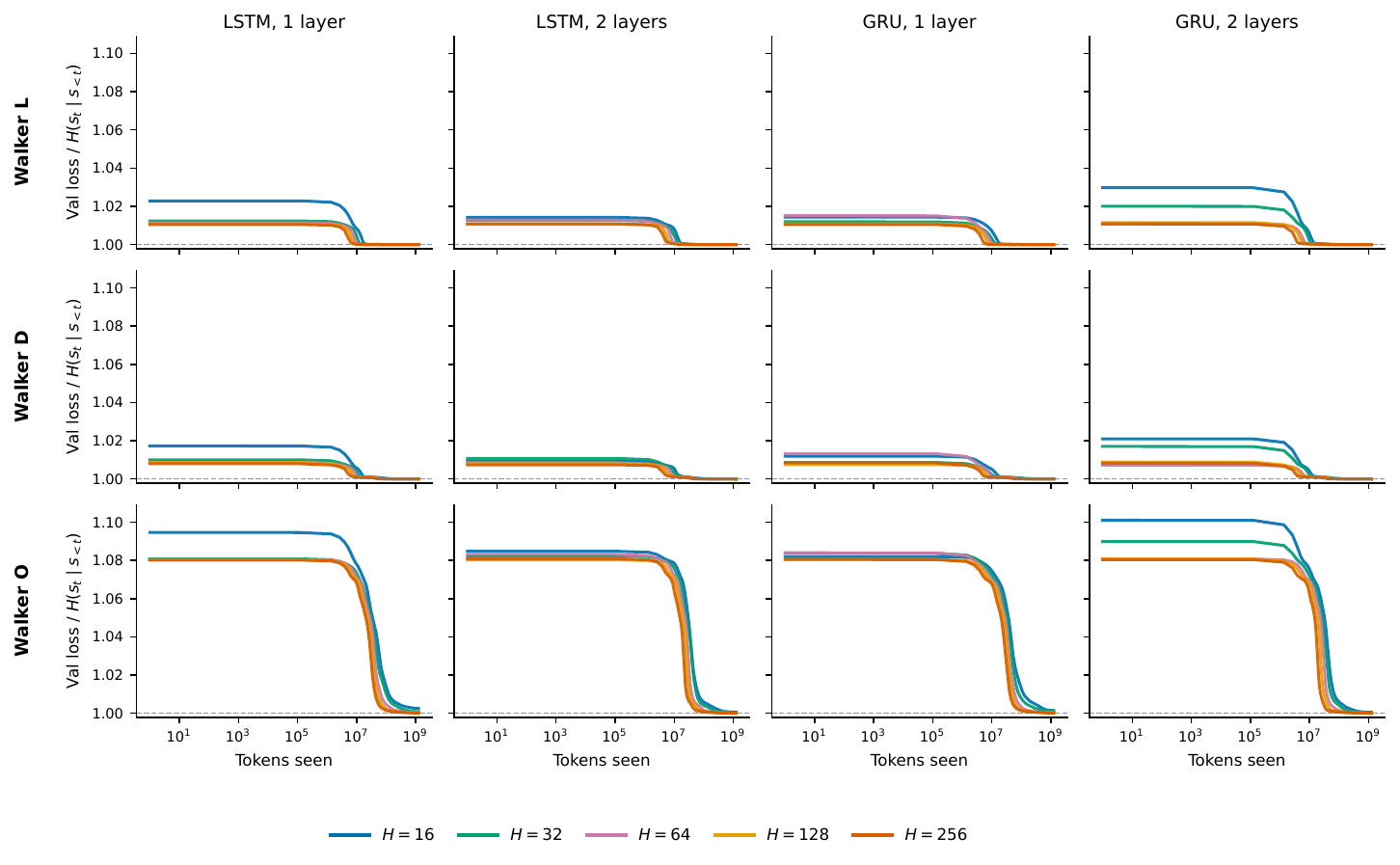}
  \caption{RNN validation loss vs.\ tokens seen for the full recurrent sweep. Rows are walker variants; columns are recurrent cell and depth. Each panel overlays hidden sizes $H\in\{16,32,64,128,256\}$. The vertical axis is the same normalized validation-loss ratio used in Tables~\ref{tab:walkers} and~\ref{tab:rnn_losses}; the dashed line marks the Bayes-optimal value $1.0$.}
  \label{fig:rnn_loss_curves_grid}
\end{figure}

Tables~\ref{tab:walkers} and~\ref{tab:rnn_losses} report final validation losses for the transformer and RNN sweeps, normalized by the exact per-position conditional entropy of the ground-truth process. The conditional entropy $H(c_{t+1}\mid s_t)$ is the irreducible uncertainty about the next token given the prefix under the true generative process: it is the cross-entropy attained by a predictor that outputs the ground-truth conditional distribution, and no causal model can do better. We refer to this floor as the \emph{Bayes-optimal} value, so that a dimensionless ratio of $1.000$ corresponds to matching the optimal predictor. Figures~\ref{fig:loss_curves_grid} and~\ref{fig:rnn_loss_curves_grid} show the corresponding validation curves over training. The transformer sweep reaches within $1.5\!\times\!10^{-4}$ of the optimum in every case; the RNN sweep follows the same convergence pattern, with the largest residual excess in the smallest Walker~O models.

\begin{table}[t]
  \centering
  \small
  \begin{tabular}{ccc|ccc}
    \toprule
    $L$ & $H$ & $d_{\rm head}$ & walker L & walker D & walker O \\
    \midrule
    1 & 1 & 32 & 1.000011 & 1.000006 & 1.000138 \\
    1 & 1 & 64 & 1.000010 & 1.000004 & 1.000044 \\
    1 & 1 & 128 & 1.000012 & 1.000005 & 1.000033 \\
    1 & 2 & 32 & 1.000010 & 1.000004 & 1.000042 \\
    1 & 2 & 64 & 1.000012 & 1.000005 & 1.000035 \\
    1 & 2 & 128 & 1.000011 & 1.000005 & 1.000038 \\
    2 & 1 & 32 & 1.000011 & 1.000006 & 1.000036 \\
    2 & 1 & 64 & 1.000011 & 1.000005 & 1.000023 \\
    2 & 1 & 128 & 1.000011 & 1.000005 & 1.000021 \\
    2 & 2 & 32 & 1.000011 & 1.000005 & 1.000023 \\
    2 & 2 & 64 & 1.000011 & 1.000004 & 1.000024 \\
    2 & 2 & 128 & 1.000011 & 1.000004 & 1.000019 \\
    \bottomrule
  \end{tabular}
  \caption{Final validation loss for every trained model, reported as the ratio $\mathrm{CE}_{\rm model} / H(c_{t+1} \mid s_t)$ averaged over context positions. The denominator is the exact per-position conditional entropy of the ground-truth process, so a value of $1.000$ corresponds to a Bayes-optimal predictor. Each column is one of the three walker variants (Section~\ref{sec:data_gen}); each row is one architecture from the appendix sweep ($L\in\{1,2\}$ layers, $H\in\{1,2\}$ heads, $d_{\rm head}\in\{32,64,128\}$).}
  \label{tab:walkers}
\end{table}

\begin{table}[t]
  \centering
  \small
  \begin{tabular}{lcc|ccc}
    \toprule
    cell & $L_{\rm rnn}$ & $H_{\rm hid}$ & walker L & walker D & walker O \\
    \midrule
    LSTM & 1 & 16 & 1.000008 & 1.000021 & 1.002547 \\
    LSTM & 1 & 32 & 1.000005 & 1.000007 & 1.000892 \\
    LSTM & 1 & 64 & 1.000004 & 1.000006 & 1.000376 \\
    LSTM & 1 & 128 & 1.000005 & 1.000006 & 1.000190 \\
    LSTM & 1 & 256 & 1.000006 & 1.000006 & 1.000130 \\
    LSTM & 2 & 16 & 1.000006 & 1.000010 & 1.000625 \\
    LSTM & 2 & 32 & 1.000005 & 1.000004 & 1.000257 \\
    LSTM & 2 & 64 & 1.000006 & 1.000003 & 1.000113 \\
    LSTM & 2 & 128 & 1.000007 & 1.000004 & 1.000075 \\
    LSTM & 2 & 256 & 1.000007 & 1.000004 & 1.000050 \\
    GRU & 1 & 16 & 1.000007 & 1.000019 & 1.001497 \\
    GRU & 1 & 32 & 1.000005 & 1.000012 & 1.001017 \\
    GRU & 1 & 64 & 1.000005 & 1.000007 & 1.000380 \\
    GRU & 1 & 128 & 1.000005 & 1.000006 & 1.000233 \\
    GRU & 1 & 256 & 1.000006 & 1.000006 & 1.000134 \\
    GRU & 2 & 16 & 1.000008 & 1.000012 & 1.000567 \\
    GRU & 2 & 32 & 1.000005 & 1.000006 & 1.000380 \\
    GRU & 2 & 64 & 1.000006 & 1.000004 & 1.000111 \\
    GRU & 2 & 128 & 1.000007 & 1.000004 & 1.000072 \\
    GRU & 2 & 256 & 1.000007 & 1.000004 & 1.000065 \\
    \bottomrule
  \end{tabular}
  \caption{Final validation loss for every trained RNN, reported as the normalized ratio $\mathrm{CE}_{\rm model}/H(c_{t+1}\mid s_t)$. Rows enumerate recurrent cell type, recurrent depth, and hidden size; columns are the three walker variants.}
  \label{tab:rnn_losses}
\end{table}

\end{document}